\UseRawInputEncoding
\documentclass{article}

\usepackage[final]{neurips2022}

\usepackage[utf8]{inputenc} % allow utf-8 input
\usepackage[T1]{fontenc}    % use 8-bit T1 fonts
\usepackage{url}            % simple URL typesetting
\usepackage{booktabs}       % professional-quality tables
\usepackage{amsfonts}       % blackboard math symbols
\usepackage{nicefrac}       % compact symbols for 1/2, etc.
\usepackage{microtype}      % microtypography
\usepackage{xcolor}         % colors
\usepackage{graphicx}
\usepackage{multirow}
\usepackage{amsmath}        % aligned equations
\usepackage{bbm}
\usepackage{mathrsfs}
\usepackage{wrapfig}
\usepackage[toc,page]{appendix}
\usepackage{hyperref}
\hypersetup{
  colorlinks=true,
  linkcolor=red,
  citecolor=blue,
  filecolor=magenta,
  urlcolor=blue,
  linktocpage
}

\newcommand\nnfootnote[1]{%
  \begin{NoHyper}
  \renewcommand\thefootnote{}\footnote{#1}%
  \addtocounter{footnote}{-1}%
  \end{NoHyper}
}

\title{TarGF: Learning Target Gradient Field to Rearrange Objects without Explicit Goal Specification}

\author{
    Mingdong Wu\textsuperscript{* \rm 1, 3}, Fangwei Zhong\textsuperscript{* \rm 2, 3}, Yulong Xia\textsuperscript{\rm 1}, Hao Dong\textsuperscript{\rm  1, 4}
    \\
  $^1$ Center on Frontiers of Computing Studies, School of Computer Science, Peking University\\
  $^2$ School of Intelligence Science and Technology, Peking University\\
  $^3$ Beijing Institute for General Artificial Intelligence (BIGAI) \\
  $^4$ Peng Cheng Laboratory \\
  \texttt{\{wmingd, zfw, hao.dong\}@pku.edu.cn} \\
}

\begin{document}

\maketitle

% notations
\UseRawInputEncoding
% Domain relevant
\newcommand{\sset}{\mathcal{S}}
\newcommand{\aset}{\mathcal{A}}
\newcommand{\trans}{\mathcal{P}}
\newcommand{\B}{\mathcal{B}}
\newcommand{\init}{\rho_0}
\newcommand{\E}{\mathbb{E}}
\newcommand{\R}{\mathbb{R}}
\newcommand{\hS}{\mathbb{S}}
\newcommand{\hP}{\mathbb{P}}
\newcommand{\M}{\mathcal{M}}
\newcommand{\Z}{\mathcal{Z}}
\newcommand{\bs}{\mathbf{s}}
\newcommand{\N}{\mathcal{N}}
\newcommand{\G}{\mathcal{G}}
\newcommand{\D}{\mathcal{D}}
\newcommand{\C}{\mathcal{C}}
\newcommand{\x}{\mathbf{x}}
\newcommand{\bc}{\mathbf{c}}
\newcommand{\p}{\mathbf{p}}
\newcommand{\z}{\mathbf{z}}
\newcommand{\ac}{\mathbf{a}}
\newcommand{\e}{\mathbf{e}}
\newcommand{\s}{\mathbf{s}}
\newcommand{\vel}{\mathbf{v}}
\newcommand{\g}{\mathbf{g}}
\newcommand{\grad}{\mathcal{G}}
\newcommand{\gtar}{\g_{tar}}
\newcommand{\gsup}{\g_{sup}}
\newcommand{\gdual}{\g_{dual}}
\newcommand{\tar}{p_{tar}}
\newcommand{\safe}{p_{safe}}
\newcommand{\tarD}{S^*}
\newcommand{\supD}{S^*_{sup}}
\newcommand{\ifunc}{\mathbbm{1}}
\newcommand{\tarS}{\mathbf{\Phi}_{tar}}
\newcommand{\supS}{\mathbf{S}_{sup}}
\newcommand{\dualS}{\mathbf{S}_{dual}}
\newcommand{\fprox}{\mathbf{F}_{proxy}}
\newcommand{\fcol}{\mathbf{F}_{col}}
\newcommand{\col}{\mathbf{c}}
\newcommand{\bt}{\mathbf{t}}
\newcommand{\policy}{\pi_{\theta}}
\newcommand{\planner}{\phi}
\newcommand{\oldpolicy}{\pi_{\theta_{old}}}
\newcommand{\newpolicy}{\pi_{\theta_{new}}}
\newcommand{\score}{\mathbf{S}}
\newcommand{\method}{DualGF}

\def\eg{\emph{e.g}.} \def\Eg{\emph{E.g}.}
\def\ie{\emph{i.e}.} \def\Ie{\emph{I.e}.}
\def\cf{\emph{c.f}.} \def\Cf{\emph{C.f}.}
\def\etc{\emph{etc}.} \def\vs{\emph{vs}.}
\def\wrt{w.r.t. } \def\dof{d.o.f. }
\def\etal{\emph{et al}. }

% formulation: example-based arrangement

\vspace{-0.3cm}
\begin{abstract}
\vspace{-0.2cm}
Object Rearrangement is to move objects from an initial state to a goal state.
Here, we focus on a more practical setting in object rearrangement, \ie, rearranging objects from shuffled layouts to a normative target distribution without explicit goal specification.
However, it remains challenging for AI agents, as it is hard to describe the target distribution (goal specification) for reward engineering or collect expert trajectories as demonstrations.
Hence, it is infeasible to directly employ reinforcement learning or imitation learning algorithms to address the task.
This paper aims to search for a policy only with a set of examples from a target distribution instead of a handcrafted reward function.
We employ the score-matching objective to train a \textit{\textbf{Tar}get \textbf{G}radient \textbf{F}ield (TarGF)}, indicating a direction on each object to increase the likelihood of the target distribution.
For object rearrangement, the TarGF can be used in two ways: 1) For model-based planning, we can cast the target gradient into a reference control and output actions with a distributed path planner; 2) For model-free reinforcement learning, the TarGF is not only used for estimating the likelihood-change as a reward but also provides suggested actions in residual policy learning. 
Experimental results in ball rearrangement and room rearrangement demonstrate that our method significantly outperforms the state-of-the-art methods in the quality of the terminal state, the efficiency of the control process, and scalability. The code and demo videos are on  \url{https://sites.google.com/view/targf}.
\nnfootnote{* indicates equal contribution}

\end{abstract}

\vspace{-0.3cm}
\section{Introduction}
\vspace{-0.3cm}
\label{sec:introduction}
%%%%%%%%%%%%%%%%%%%%%%%%%%%%%%%%%%%%%%%%%%%%%%%%%%%%%%%%%%%%%%%%%%%%%%%%%%%%%%%
% Problem
%%%%%%%%%%%%%%%%%%%%%%%%%%%%%%%%%%%%%%%%%%%%%%%%%%%%%%%%%%%%%%%%%%%%%%%%%%%%%%%
% 场面话
\begin{wrapfigure}{tr}{0.48\textwidth}
\centering
\vspace{-15pt}
\includegraphics[width=0.48\textwidth]{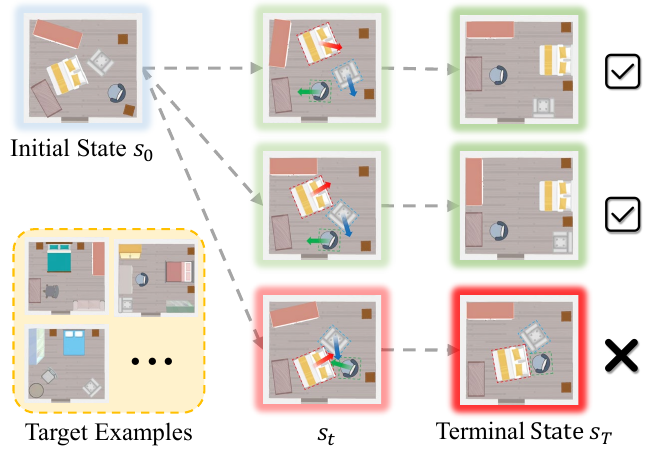}
\caption{Our goal: Learning to rearrange objects \textbf{without explicit goal specification}.}
\label{teaser}
\vspace{-20pt}
\end{wrapfigure}

As shown in Fig.~\ref{teaser}, we consider object rearrangement \emph{without explicit goal specification}~\cite{ICML12Arrange, abdo2015robot}, where the agent is required to manipulate a set of objects from an initial layout to a normative distribution. 
This task is taken for granted by humans~\cite{ricciuti1965object} and widely exists in our daily life, such as tidying up a table~\cite{neatnet}, placing furniture~\cite{wang2020scene}
, and sorting parcels~\cite{kim2020sortation}.

However, compared with conventional robot tasks~\cite{yu2020meta, zhu2017target, luo2019end, kiran2021deep, zuo2019craves}, there are two critical challenges in our setting:
1) Hard to clearly define the target state (goal), as the target distribution is diverse and the patterns are difficult to describe in the program, \eg, how to evaluate the \textit{tidiness} quantitatively? 
As a result, it is difficult to manually design a reward/objective function, which is essential to modern control methods, \eg, deep reinforcement learning (DRL)~\cite{mnih2015human}.
2) In the physical world, the agents also need to find an executable path to reach the target state efficiently and safely, \ie, finding a short yet collision-free path to reach the target distribution.
To this end, the agents are required to jointly learn to evaluate the quality of the state (reward) and move objects to increase the quality efficiently. In other words, the agents should answer the two questions: \textit{What kind of layouts does the user prefer}, and \textit{How to manipulate objects to meet the target distribution}?

Thus, searching for a policy without dependence on the explicit goal state and shaped reward is necessary.
Imitation learning (IL) can learn a reward function~\cite{ho2016generative, SQIL2019, EBIL} or learn a policy~\cite{BC1, BC2,BC3,zhong2021towards} from the expert's supervision, \eg, demonstrating expected trajectories. However, collecting a large number of expert trajectories is expensive.
Recently, example-based reinforcement learning (RL)~\cite{VICE2018, RCE} has tried to reduce the reliance on expert trajectories, \ie, learning a policy only with examples of successful outcomes.
However, it is still difficult to learn a classifier to provide accurate reward signals for learning in the case of high-dimensional state and action space, as there are many objects to control in object rearrangements. 

%%%%%%%%%%%%%%%%%%%%%%%%%%%%%%%%%%%%%%%%%%%%%%%%%%%%%%%%%%%%%%%%%%%%%%%%%%%%%%%
% METHOD
%%%%%%%%%%%%%%%%%%%%%%%%%%%%%%%%%%%%%%%%%%%%%%%%%%%%%%%%%%%%%%%%%%%%%%%%%%%%%%%
%%% Overview %%%
% Inspiration %
In this work, we separate the task into two stages: 1) learning a score function to estimate the target-likelihood-change between adjacent states, and 2) learning/building a policy to rearrange objects with the score function.
Inspired by the recent advances in score-based generative models~\cite{VanillaScoreMatching, SDEScoreMatching, 21Molecule,LST,SDEdit}, we employ the denoising score-matching objective~\cite{denosingScoreMatching} to train a  \textit{target score network} for estimating the \textit{target gradient field}, \ie, the gradient of the log density of the target distribution.
The target gradient field can be viewed as object-wise guidance that provides a pseudo velocity for each object, pointing to regions with a higher density of the target distribution, \eg, tidier  configurations. 
To rearrange objects in the physical world, we demonstrate two ways to leverage the target gradient field in control.
1) We integrate the gradient-based guidance with a model-based path planner to rearrange objects without collision.
2) We further derive a reward function for reinforcement learning by estimating the likelihood change via the target gradient.
With the TarGF-based reward and a TarGF-guided residual policy network~\cite{Residual}, we can train an effective RL-based policy.

%%%%%%%%%%%%%%%%%%%%%%%%%%%%%%%%%%%%%%%%%%%%%%%%%%%%%%%%%%%%%%%%%%%%%%%%%%%%%%%
% Experiments & Summary
%%%%%%%%%%%%%%%%%%%%%%%%%%%%%%%%%%%%%%%%%%%%%%%%%%%%%%%%%%%%%%%%%%%%%%%%%%%%%%%
% 实验
In experiments, we evaluate the effectiveness of our methods on two typical rearrangement tasks: \textit{Ball Rearrangement} and \textit{Room Rearrangement}. 
First, we introduce three metrics to analyse the quality of the terminal state and the efficiency of the control process.
In \textit{Ball Rearrangement}, we demonstrate that our control methods can efficiently reach a high-quality and diverse terminal configuration. 
We also observe that our methods can find a path with fewer collisions and shorter lengths to converge compared with learning-based baselines~\cite{SQIL2019, RCE} and planning-based baselines~\cite{ORCA}.
In \textit{Room Rearrangement}, we further evaluate our methods' feasibility in complex scenes with multiple heterogeneous objects.

% Contributions
Our contributions are summarised as follows:
\begin{itemize}
    \vspace{-0.2cm}
    \item We introduce a novel score-based framework that estimates a \textit{target gradient field (TarGF)} via denoising score matching for object rearrangement.
    \item We point out two usages of the target gradient field in object rearrangement: 1) providing guidance in moving direction; 2) deriving a reward for learning.  
    \item We conduct experiments to demonstrate that both traditional planning and RL methods can benefit from the target gradient field and significantly outperform the state-of-the-art methods in sample diversity, realism and safety.
\end{itemize}
\vspace{-0.3cm}

\section{Related Works}
\vspace{-0.2cm}
\subsection{Object and Scene (Re-)Arrangement}
\vspace{-0.2cm}
% Most relevant to our experiments.
The object (re-)arrangement is a long-studied problem in robotics community~\cite{abdo2015robot, schuster2012learning, abdo2016organizing} and the graphics community~\cite{fisher2012example, MakeItHome2011, RJMCMC2012, SceneNet2016, songchun2018}. However, most of them focus on manually designing rules/ energy functions to find a goal~\cite{abdo2015robot, schuster2012learning, abdo2016organizing} or synthesising a scene configuration~\cite{MakeItHome2011, RJMCMC2012, SceneNet2016, songchun2018} that satisfies the preference of the user. 
The most recent work~\cite{neatnet} tries to learn a GNN to output an arrangement tailored to human preferences. However, they neglect the physical process of rearrangement.
Hence, the accessibility and transition costs from the initial and target states are not guaranteed.
Considering the physical interaction, the recent works on scene rearrangement~\cite{rearrange, visualrearrange} mitigate the goal specification problem by providing \textit{a goal scene} to the agent, making the reward shaping feasible.
Concurrent works~\cite{ECCV22HouseKeep, ECCV22TIDEE} also notice the necessity of automatic goal inference for tying rooms and exploit the commonsense knowledge from Large Language Model (LLM) or memex graph to infer rearrangements goals when the goal is unspecified.
Another concurrent work~\cite{ICML22Arrangement} aims at discovering generalisable spatial goal representations via graph-based active reward learning for 2D object rearrangement.
In this paper, we focus on a more practical setting of rearrangement: \emph{how to estimate the similarity (i.e., the target likelihood) between the current state and the example sets and manipulate objects to maximise it}. Here, only a set of positive target examples are required to learn the gradient field rather than prior knowledge about specific scenarios.

\subsection{Learning without Reward Engineering} % Learning without Reward Engineering
% Imitation Learning  vs.  Inverse RL
% trajectory vs. 
Without hand-engineered reward, previous works explore algorithms to learn a control policy from expert demonstrations via behavioural cloning~\cite{BC1, BC2, BC3, ho2016generative} or inverse reinforcement learning~\cite{AIRL, EBIL, DAC, MaxEntIRL}.
Imitation learning (IL)~\cite{BC1, BC2, BC3} aims to directly learn a policy by cloning the behaviour from the expert trajectories. 
The inverse reinforcement learning (IRL)~\cite{AIRL, EBIL, DAC, MaxEntIRL} tries to learn a reward function from data for subsequent reinforcement learning.
Considering the difficulty in collecting expert trajectories, some example-based methods are proposed to learn a policy only with a set of successful examples~\cite{VICE2018, SQIL2019, RCE}.
The most recent work is RCE~\cite{RCE} which directly learns a recursive classifier from successful outcomes and interactions for policy optimisation.
We also try to develop an example-based control method for rearrangement with a set of examples. 
Our method can learn the target gradient field from the examples without any additional effort in reward engineering.
The target gradient field can provide meaningful reward signals and action guidance for reinforcement learning and traditional path planners. To the best of our knowledge, such usage of gradient field is not explored in previous works.

\section{Problem Statement}
\label{sec:problem statement}
We aim to learn a policy that moves objects to reach states close to a target distribution without relying on reward engineering. 
Similar to the \textit{example-based control}~\cite{RCE}, the grounded \textit{target distribution} $\tar(\s)$ is unknown to the agent, and the agent is only given a set of \textit{target examples} $\tarD = \{\s^*\}$ where $ \s^* \sim \tar(\s)$.
In practice, these examples can be provided by the user without accessing the dynamics of the robot. 
The agent starts from an initial state $\s_0 \sim p_0(\s_0)$, where $N$ objects are randomly placed in the environment. 
At time step $t$, the agent takes action $\pi(\ac_t|\s_t)$ imposed on the objects to reach the next state with dynamics $p(\s_{t+1} | \s_t, \ac_t)$.
Here, the goal of the agent is to search for a policy $\pi^*(\ac_t|\s_t)$
that maximises the \textit{discounted log-target-likelihood} of the future states:
\begin{equation}
    \pi^* = \mathop{\arg\max}\limits_{\pi}E_{ \rho(\s_0), \tau \sim \pi} \left[ \sum\limits_{\s_t \in \tau}  \gamma^t \log \tar(\s_t) \right]
\label{eq:arrangement objective}
\end{equation}
where $\gamma \in (0, 1]$ denotes the discount factor. 
Notably, we assume that $\tar(\s) > 0$ everywhere since we can perturb the original data with a tiny noise (\eg, $\N(0, 0.00001)$) to ensure the perturbed density is always positive. \cite{VanillaScoreMatching} also used this trick to tackle the manifold hypothesis issue.
This objective reveals three challenges:
\textbf{1) Inaccessibility problem:} The grounded target distribution $\tar$ in Eq.~\ref{eq:arrangement objective} is inaccessible. Thus, we need to learn a function to approximate the \textit{log-target-likelihood} for policy search. \textbf{2) Sparsity problem:} The log-target-likelihood is sparse in the state space due to the \textit{Manifold Hypothesis} mentioned in~\cite{VanillaScoreMatching}. Hence, even if we have access to the target distribution $\tar$, it is still difficult to explore the high-density region.
\textbf{3) Adaptation problem:} The dynamics $p(\s_{t+1} | \s_t, \ac_t)$ is also inaccessible. 
To maximise the cumulative sum in Eq.~\ref{eq:arrangement objective}, the agent is required to adapt to the dynamics to efficiently increase the target likelihood. 

To address these problems, we partition the task into 1) learning to estimate the log-target-likelihood (reward) of the state and 2) learning/building a policy to rearrange objects to adapt to the dynamics.

\section{Method}
%%%%%%%%%%%%%%%%%%%%%%%%%%%%%%%%% Method Illustration %%%%%%%%%%%%%%%%%%%%%%%%%%%%%%%%
\begin{figure}[!tb]
    \centering
    \vspace{-1mm}    

    \label{fig:method}
    \resizebox{\columnwidth}{!}{
    \scalebox{1}{
        \includegraphics[width=\columnwidth]{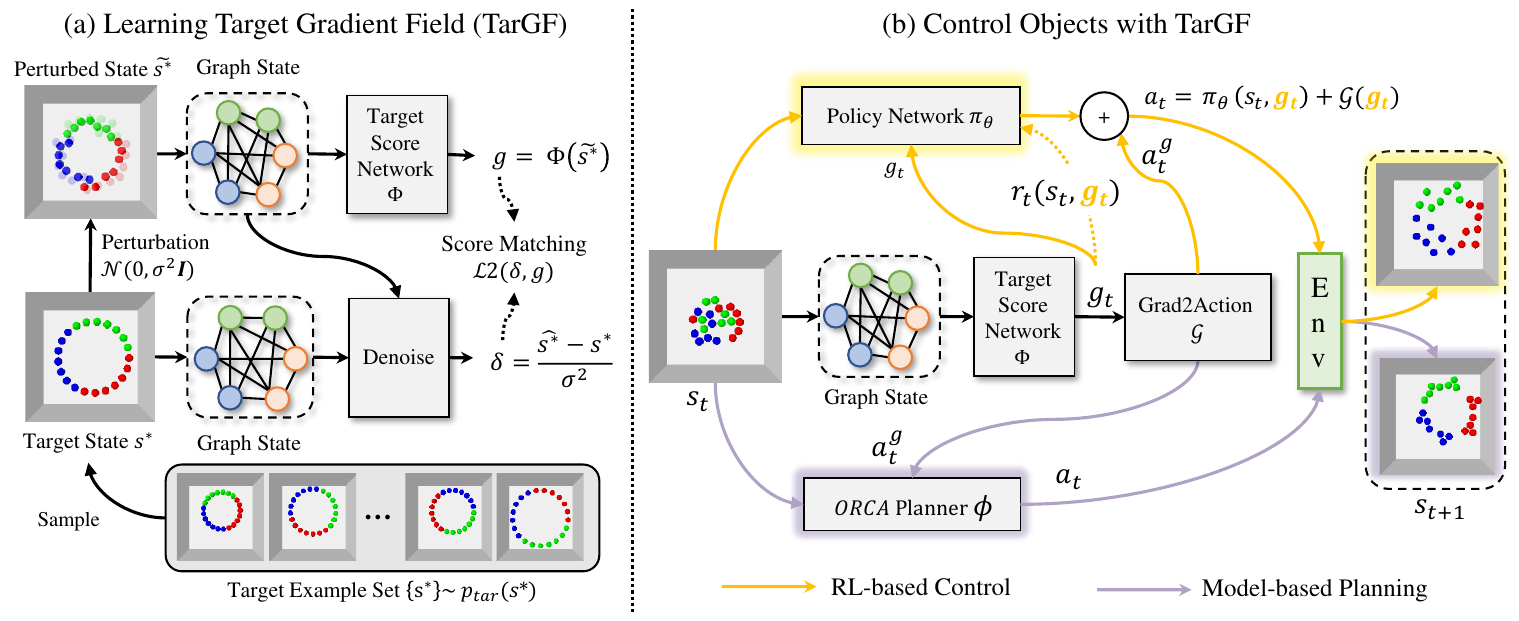}
    }}
    \caption{An overview of our method. (a) We train the target score network via score matching. The target examples are first perturbed by Gaussian noise. The target score network is forced to match the denoising direction of the perturbation.  (b) Our framework is based on the trained target score network. The target score network provides exploration guidance and reward estimation for RL in the model-free setting. In the model-based setting, the TarGF provides reference velocities for a model-based planner based on ORCA. The planner then outputs collision-free velocities for objects.}
    \vspace{-1mm}    
\end{figure}
%%%%%%%%%%%%%%%%%%%%%%%%%%%%%%%%%%%%%%%%%%%%%%%%%%%%%%%%%%%%%%%%%%%%%%%%%%%%%%%%%%%%%%%%

\subsection{Motivation}
\label{sec: motivation}
Estimating the \textit{gradient} of the log-target-likelihood $\nabla_{\s} \log \tar(\s)$ can tackle the first two problems mentioned in Sec.~\ref{sec:problem statement}:
For the \textit{inaccessibility} problem, we can approximate the likelihood increment between two adjacent states by the first-order Taylor expansion $\log\tar(\s_{t+1}) - \log\tar(\s_{t}) \approx \langle \nabla_{\s} \log \tar(\s_{t}), \s_{t+1} - \s_{t}\rangle$.
This helps to derive a surrogate objective~(\ie, Eq.~\ref{eq: delta likelihood estimation-3}) of Eq.~\ref{eq:arrangement objective}.

For the \textit{sparsity} problem, we can leverage the gradient field to help with exploration since the gradient $\nabla_{\s} \log \tar(\s)$ indicates the fastest direction to increase the target likelihood.
To estimate $\nabla_{\s} \log \tar(\s)$, we employ the score-matching~\cite{SDEScoreMatching}, a novel generative model that achieved impressive results in many research areas recently\cite{21Molecule, SDEdit, LST}, to train a \textit{target score network} $\tarS: \sset \rightarrow \sset$.

To address the \textit{adaptation} problem, we further demonstrate two approaches to incorporate the trained target gradient field $\tarS$ with control algorithms for object rearrangement:
1) For the model-based setting, our framework casts the target gradient into a reference control and outputs an action with a distributed path planner.
2) For the model-free setting (reinforcement learning), we leverage the target gradient to estimate reward and provide suggested action in residual policy learning~\cite{Residual}. 

In the following, we will first introduce how to train the target gradient field $\tarS$ via score-matching in Sec.~\ref{sec:gradient_field} and then introduce our rearrangement framework under the model-based and model-free settings in Sec.~\ref{sec:planning} and Sec.~\ref{sec:policy} respectively.

\subsection{Learning the Target Gradient Field from Examples}
\label{sec:gradient_field}
The target score network $\tarS$ is the \textbf{core module} of our framework, which aims at estimating the \textit{score function}~(\ie the gradient of log-density) of the target distribution $\nabla_{\s}\log\tar(\s)$. 

To train the target score network, we adopt the Denoising Score-Matching (DSM) objective proposed  by~\cite{denosingScoreMatching}, which can guarantee a reasonable estimation of the $\nabla_{\s} \log \tar(\s)$.
In training, we pre-specify a noise distribution $q_{\sigma}(\widetilde \s|\s) = \N(\widetilde \s; \s, \sigma^2I)$. Then, DSM matches the output of the target score network with the score of the perturbed target distribution $q_{\sigma}(\widetilde \s) = \int q_{\sigma}(\widetilde \s|\s) \tar(\s) d\s$:
\begin{equation}
    \frac{1}{2} \E_{q_{\sigma}(\widetilde \s|\s), \tar(\s)}\left[||\tarS(\widetilde \s) - \nabla_{\widetilde \s}\log q_{\sigma}(\widetilde \s|\s)||^2_2\right] = \frac{1}{2} \E_{q_{\sigma}(\widetilde \s|\s), \tar(\s)}\left[||\tarS(\widetilde \s) - \frac{\s - \widetilde\s}{\sigma^2}||^2_2\right]
\label{eq:DSM-Gaussian}
\end{equation}
The DSM objective guarantees the optimal score network satisfies $\tarS^*(\s) = \nabla_{\s}q_{\sigma}(\s)$ almost surely. When $\sigma$ is small enough, we have $\nabla_{\s}q_{\sigma}(\s) \approx \nabla_{\s} \log \tar(\s)$, so that $\tarS^*(\s) \approx \nabla_{\s} \log \tar(\s)$.

In practice, we adopt a variant of DSM proposed by~\cite{SDEScoreMatching}, which conducts DSM under different noise scales simultaneously.
This way, we can efficiently try different noise scales after only one training.
The target score network is implemented as a graph-based network $\tarS(\s, t)$, which is conditioned on a noise level $t \in (0, 1]$.
The input state$\s = [\s^1, \s^2, ..., \s^N]$ is constructed as a fully-connected graph that takes each object's state~(\eg position, orientation) $\s^i \in \R^{f_s}$ and static properties~(\eg category, bounding-box) $\p^i \in \R^{f_p}$ as the node feature $[\s^i, \p^i] \in \R^{f_s+f_p}$.
The input graph is pre-processed by linear feature extraction layers and then passed through several Edge-Convolution layers. 
After message-passing, the output on the i-th node serves as the component of the target gradient on the i-th object's state $\tarS^i(\s) \in \R^{f_s}$. 
We defer structure details to Appendix.~\ref{sup: ours detail}.

\subsection{Model-based Planning with the Target Gradient Field}
\label{sec:planning}
The target gradient $\g = \tarS(\s)$ can be translated into a \textit{gradient-based action} $\ac^g = \grad(\g)$, where $\grad$ denotes the gradient-to-action translation.
A gradient over the state space can be viewed as velocities imposed on each object.
For instance, if the state of each object is a 2-dimensional position $\s_i = [x, y]$, then the target gradient component on the i-th object can be viewed as linear velocities on two axes $\tarS(\s)^i = [v_x, v_y]$.
When the action space is also velocities, we can construct $\ac^g$ by simply projecting the target gradient into the action space $\ac^g = \grad(\g) = \frac{v_{max}}{||\g||_2}\cdot\g$, where the $v_{max}$ denotes the speed limit.
Following the gradient-based action, objects will move towards directions that may increase the likelihood of the next state, which meets the need for the rearrangement task.

However, the gradient-based action will potentially lead objects to collide with each other since the target gradient cannot infer the environment constraints from only target examples.

To adapt the gradient-based action to the environment dynamics, we incorporate the target gradient field with a model-based off-the-shelf collision-avoidance algorithm \textit{ORCA}. 
The ORCA planner $\planner$ takes the gradient-based action $\ac^g$ as reference velocities and then outputs the collision-free velocities $\ac = \planner(\ac^g)$ with the least modification to the reference velocities. 
In this way, the adapted action can lead objects to move towards a higher likelihood region efficiently and safely.
For more details of the ORCA planner, we defer to Appendix~\ref{sec: orca_detail}.

\subsection{Learning Policy with the Target Gradient Field} 
\label{sec:policy}
The model-based approach to `correct' the target gradient assumes the objects are of circular shape.
This limits the scope of this method when objects are of non-circular shape~(\eg, furniture). To this end, we propose a \textit{model-free approach} based on reinforcement learning (RL) for object rearrangement, where the agent needs to adapt the dynamics via online interactions.

To tackle the inaccessibility problem mentioned in Sec.~\ref{sec:problem statement}, we first derive an equivalent $J^*(\pi)$ of the original objective in Eq.~\ref{eq:arrangement objective} by subtracting a constant from the original objective:
\begin{equation}
\begin{aligned}
J(\pi)\iff&\E_{ \rho(\s_0), \tau \sim \pi} [ \sum\limits_{\s_t \in \tau} \gamma^t \log \tar(\s_t) ] - \underbrace{\frac{\E_{ \rho(\s_0)} [\log \tar(\s_0)]}{1-\gamma}}_{constant \quad C} \\ 
=&\E_{ \rho(\s_0), \tau \sim \pi} [\sum\limits_{1 \leq t \leq T}\gamma^t \sum\limits_{1\leq k \leq t}[\log \tar(\s_k) - \log \tar(\s_{k-1}) ]] \overset{def}{=} J^*(\pi)
\label{eq: delta likelihood estimation-1}
\end{aligned}
\end{equation}
Further, we derive a surrogate objective $\hat{J}(\pi)$ to approximate the $J^*(\pi)$. We notice that in most physical simulated tasks, the distance between two adjacent states $||\s_{t}-\s_{t-1}||_2$ is quite small, which inspires us to approximate the log-likelihood-change of the adjacent states using the Taylor expansion:
\begin{equation}
\begin{aligned}
J^*(\pi)
\approx\E_{ \rho(\s_0), \tau \sim \pi}[\sum\limits_{1 \leq t \leq T} \gamma^t \underbrace{\sum\limits_{1\leq k \leq t}\langle \tarS(\s_{k-1}), \s_{k}-\s_{k-1} \rangle}_{r_t}] \overset{def}{=} \hat{J}(\pi)
\label{eq: delta likelihood estimation-3}
\end{aligned}
\end{equation}
In this way, we can optimise the surrogate objective $\hat{J}(\pi)$ by assigning $\sum\limits_{1\leq k \leq t}\langle \tarS(\s_t), \s_{k}-\s_{k-1} \rangle$ as the immediate reward.
In practice, we simply assign the last term of the cumulative summation as the immediate reward $ r_t = \langle \tarS(\s_t), \s_{t}-\s_{t-1} \rangle$, and this version is shown to be the most effective.

To further tackle the sparsity problem, our idea is to build a residual policy $\policy$ upon the gradient-based action $\ac_t^g$ mentioned in Sec.~\ref{sec:planning}, as $\ac_t^g$ helps to explore the regions with a high target likelihood:
\begin{equation}
    \underbrace{\ac_t = \policy(\s_t, \g_t) + \ac^g_t}_{output \, action},
    \quad \underbrace{\ac^g_t = \grad(\g) =\frac{v_{max}}{||\g_t||_2}\cdot\g_t}_{gradient-based \, action},
    \quad \underbrace{\g_t = \tarS(\s_t)}_{target \, gradient}
\label{eq: residual learning derivation}
\end{equation}
This approach is similar to the residual policy learning proposed by~\cite{Residual}. 
The $\ac^g_t$ serves as an initial policy, and the residual policy network $\policy$ takes the target gradient $\g_t$ and the current state $\s_t$ as input and outputs $\pi(\s_t, \g_t)$ to `correct' the $\ac^g_t$.
The agent finally outputs a `corrected' action $\ac_t = \pi(\s_t, \g_t) + \ac^g_t$.
In this way, the agent can benefit from the efficient exploration and have less training burden as it `stands on the giant's shoulder' (\ie the gradient-based action $\ac^g_t$).

% details 
We employ the multi-agent Soft-Actor-Critic(SAC)~\cite{SAC} as the reinforcement learning algorithm backbone. 
Each object is regarded as an agent and can communicate with all the other agents.
At each time step $t$, the reward for the i-th agent is $r_t = r^{tar}_t + \lambda * r^{i}_t$, where $r^{tar}_t = \sum\limits_{1\leq k \leq t}\langle \tarS(\s_t), \s_{k}-\s_{k-1} \rangle$ is a centralised target likelihood reward, $\lambda > 0$ is a hyperparameter and $r^{i}_t = \sum_{j \neq i} c_{i, j}$ is a decentralised collision penalty where $c_{i, j}$ is a collision penalty. Specifically, when a collision is detected between object $i$ and $j$, $c_{i, j}=-1$, otherwise $c_{i, j}=0$.

\section{Experiment Setups}
\subsection{Environment}
% observation space, action space, goal/task
We design two object rearrangement tasks without explicit goal specification for evaluation: \textit{Ball Rearrangement} and \textit{Room Rearrangement}, where the former uses controlled environments for better numerical analysis and the latter is built on a real dataset with implicit target distribution.
% to demonstrate the potential of our method on a more challenging task.

\begin{figure}[b]
    \centering
    \vspace{-5mm}
    \resizebox{0.98\columnwidth}{!}{
    \scalebox{1}{
        \includegraphics[width=\columnwidth]{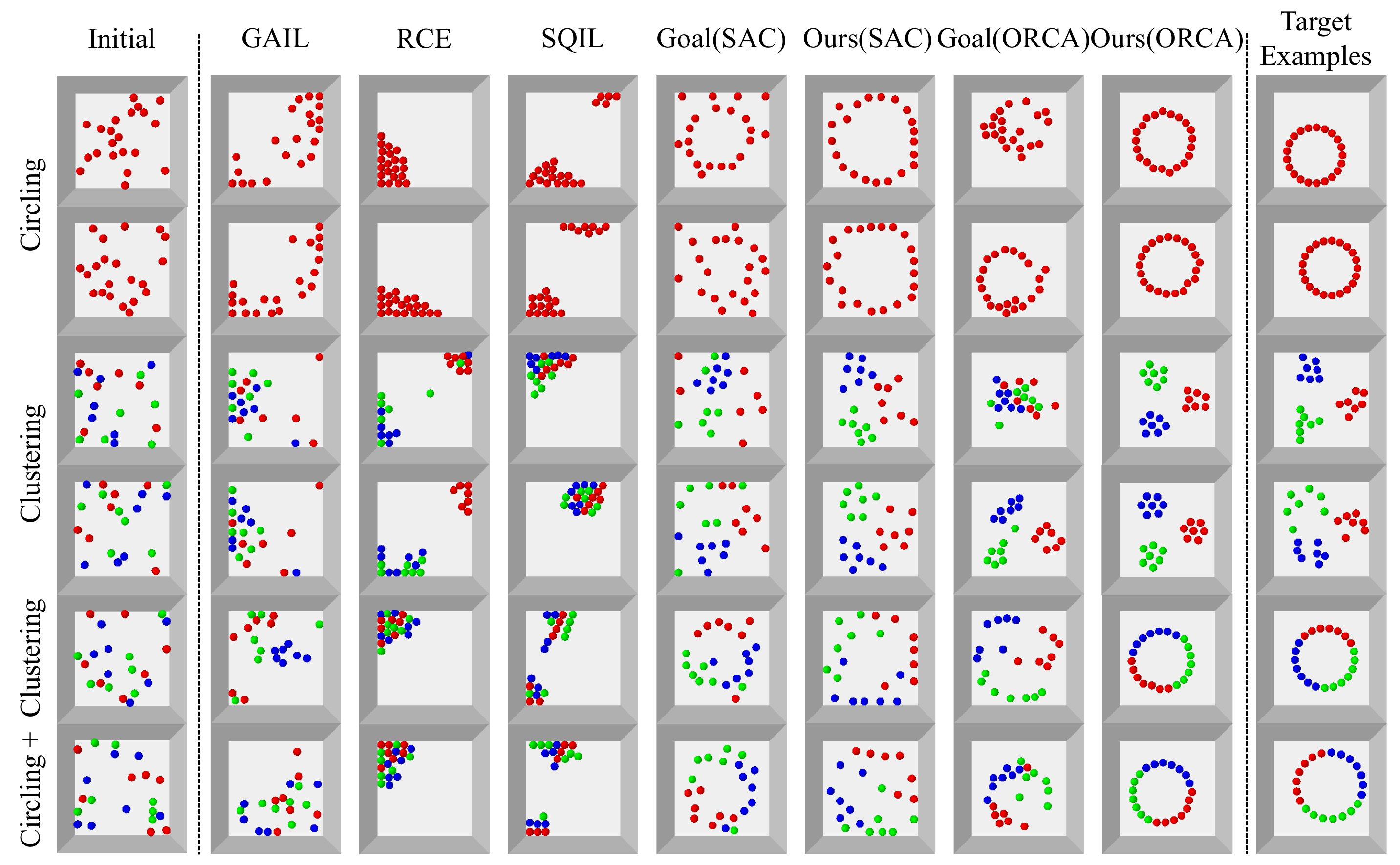}
    }}
        \caption{
    Starting from the same initial states on the far left, we demonstrate the rearrangement results of different methods.
    On the far right, we demonstrate some target examples of each task.
    }   
    \vspace{-5mm}
    \label{fig:ball qualitative}  
\end{figure}

\textbf{Ball Rearrangement} includes three environments with increasing difficulty: \textit{Circling}, \textit{Clustering}, and the hybrid of the first two, \textit{Circling + Clustering}, as shown in Fig.~\ref{fig:ball qualitative}. 
There are 21 balls of the same geometry in each task.
The balls are divided equally into three colours in \textit{Clustering} and \textit{Circling+Clustering}.
The goals of the three tasks are as follows: \textit{Circling} requires all balls to form a circle. The circle's centre can be anywhere in the environment; \textit{Clustering} requires all balls to form into three clusters by colour. The joint centre of each cluster has two choices~(\ie, red-green-blue or red-blue-green clockwise);  \textit{Circling + Clustering} requires all balls to form into a circle where balls of the same type are adjacent.
To slightly increase the complexity of the physical dynamics, we enlarge the lateral friction coefficient of each ball and make each ball with different physical properties, \eg, different friction coefficients and masses.
The target examples of each task are sampled from an explicit process. 
Thus, we can define a `pseudo likelihood' on a given state according to the sampling process to measure the similarity between the state and the target distribution. 

%%%%%%%%%%%%%%%%%%%%%%%%%%% Ball Qualitative %%%%%%%%%%%%%%%%%%%%%%%%%%%%%%%%%%%%%%%%%%%%
%%%%%%%%%%%%%%%%%%%%%%%%%%%%%%%%%%%%%%%%%%%%%%%%%%%%%%%%%%%%%%%%%%%%%%%%%%%%%%%%%%%%%%%%

\textbf{Room Rearrangement} is built on a more realistic dataset, 3D-FRONT~\cite{3dfront}. 
After cleaning the data, we get a dataset with 839 rooms, 756 for training, \eg, target examples, and 83 for testing. 
In each episode, we sample a room from the test split and shuffle the room via a Brownian Process at the beginning to get an initial state. 
The agent can assign angular and linear velocities for each object in the room at each time step.
To tidy up the room, the agent must learn implicit knowledge from the target examples, as shown in Fig.~\ref{fig:room qualitative}.
More details are in the Appendix~\ref{sec: rearrange details}.

%%%%%%%%%%%%%%%%%%%%%%%%% Room Qualitative %%%%%%%%%%%%%%%%%%%%%%%%%%%%%%%%%%%%%%%%%%%%%
\begin{figure}[t]
\centering
    \vspace{-5mm}

    \resizebox{\columnwidth}{!}{
    \scalebox{1}{
        \includegraphics[width=0.99\columnwidth]{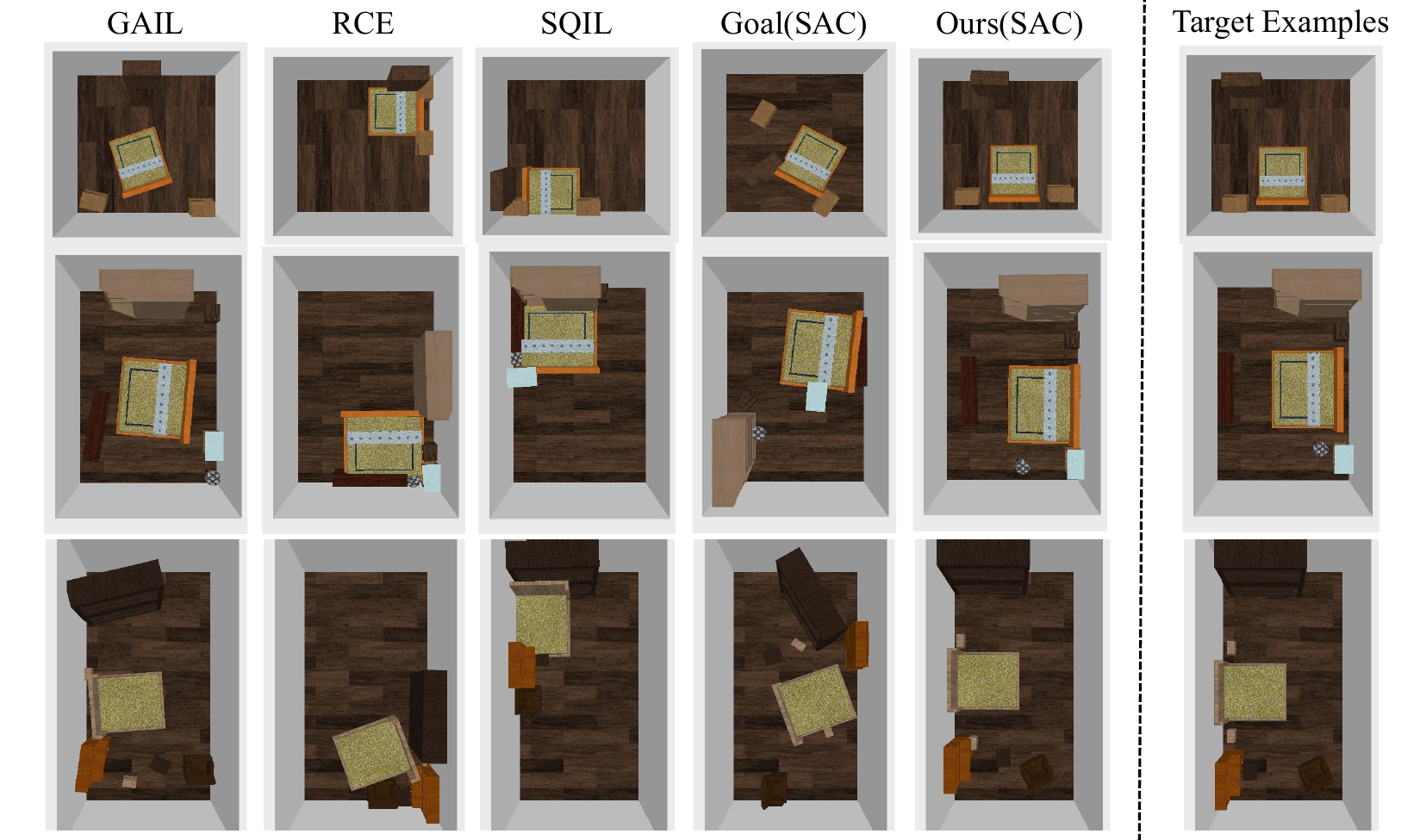}
    }}
    \vspace{-5mm}
    \caption{Qualitative results on room rearrangement. We obtain eight rearrangement results of each method starting from the same set of eight initial states. Then we demonstrate the rearrangement result closest to each method's target example.
    }
    \label{fig:room qualitative}
\end{figure}
%%%%%%%%%%%%%%%%%%%%%%%%%%%%%%%%%%%%%%%%%%%%%%%%%%%%%%%%%%%%%%%%%%%%%%%%%%%%%%%%%%%%%%%%

\subsection{Evaluation}
For baseline comparison and ablation studies, we collect a fixed number of trajectories $T = \{ \tau_i \}$ starting from the same initial states for each of the 5 random seeds.
For each random seed, we collect 100 and $83 \times 8$~(8 initial states for each room) trajectories for ball and room rearrangement, respectively.
Then we calculate the following metrics on the trajectories, where the PL is only reported in ball rearrangement. More details of evaluation are in the Appendix~\ref{sec: eval_details}.  

\textbf{Pseudo-Likelihood (PL)} 
measures the efficiency of the rearrangement process and the sample quality of the results.
We specify a \textit{pseudo-likelihood function} $\fprox: \sset \rightarrow \R^+$ for each environment that measures the similarity between a given state and the target examples.
For each time step $t$, we report the mean PL over trajectories $\E_{\tau \sim T}[\fprox (\s_t)]$ and the confidence interval over 5 random seeds.
We do not report PL on room rearrangement as it is hard to program human preferences.

\textbf{Coverage Score (CS)} measures the diversity and fidelity of the rearrangement results (\ie the terminal states $S_{T}=\{\s_T \}$ of the trajectories) by calculating the Minimal-Matching-Distance~(MMD)~\cite{MMD} between $S_{T}$ and a fixed set of examples $S_{gt}$ from $\tar$: $\sum\limits_{\s_{gt} \in S_{gt}} \min\limits_{\s_T \in S_{T}} ||\s_{gt} - \s_T||$.
If the rearrangement results of a method miss some modes in the ground truth example set, the coverage score will increase to hurt the performance.

\textbf{Averaged Collision Number (ACN)} %partially 
reflects the safety and efficiency of the rearrangement process since object collision will lead to object blocking and deceleration.
ACN is calculated as  $\sum_{\tau \in T}\sum_{\s_t \in \tau} \col_t$, where $\col_t$ denotes the total collision number at time step $t$.

\subsection{Baselines} 

We compare our framework, \ie, \textit{Ours~(ORCA) and Ours~(SAC)}, with planning-based baselines and learning-based baselines.
Besides, we design heuristic baselines as the upper bounds for ball rearrangement tasks.
All the reinforcement learning-based methods share the same network design, capacity and hyperparameters.
More details of baselines are in Appendix.~\ref{sup: baselines}.

\textbf{Leaning-based Baselines:} \textit{GAIL:} A classical inverse RL method that trains a discriminator as the reward function. \textit{RCE:} A strong example-based reinforcement learning method proposed recently. \textit{SQIL:} A competitive example-based imitation learning method. \textit{Goal-RL:} An `open-loop' method that first generates a goal at the beginning of the episode via a VAE trained from the target examples and then trains a goal-conditioned policy to reach the proposed goal via RL.

\textbf{Planning-based Baselines:} \textit{Goal~(ORCA):} 
This method first generates a goal via the same VAE as in \textit{Goal~(SAC)}. Then the agent assigns an action pointing to the goal as reference control and adjusts the preference velocity using ORCA, similar to Sec.~\ref{sec:planning}.

\textbf{Heuristic-based Baseline:} \textit{Oracle:} We slightly perturb the samples from the target distribution and take these samples as the rearrangement results. 
This method's rearrangement results will be close to the target distribution yet different from the target examples.
Hence, we use this method to normalise the PL curve and CS bars.

%%%%%%%%%%%%%%% Comparative Results: baseline+ablation %%%%%%%%%%%%%%%%%%%%%%
\begin{figure}[tb]
\centering
    \vspace{-10mm}    
    \includegraphics[width=\linewidth]{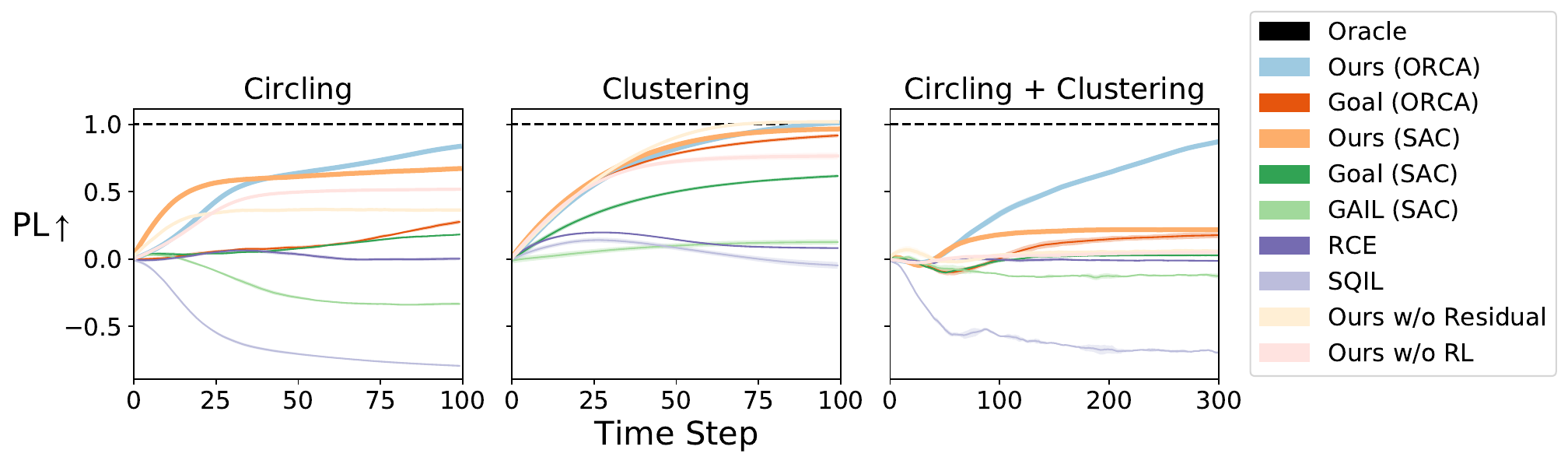}
    \includegraphics[width=\linewidth]{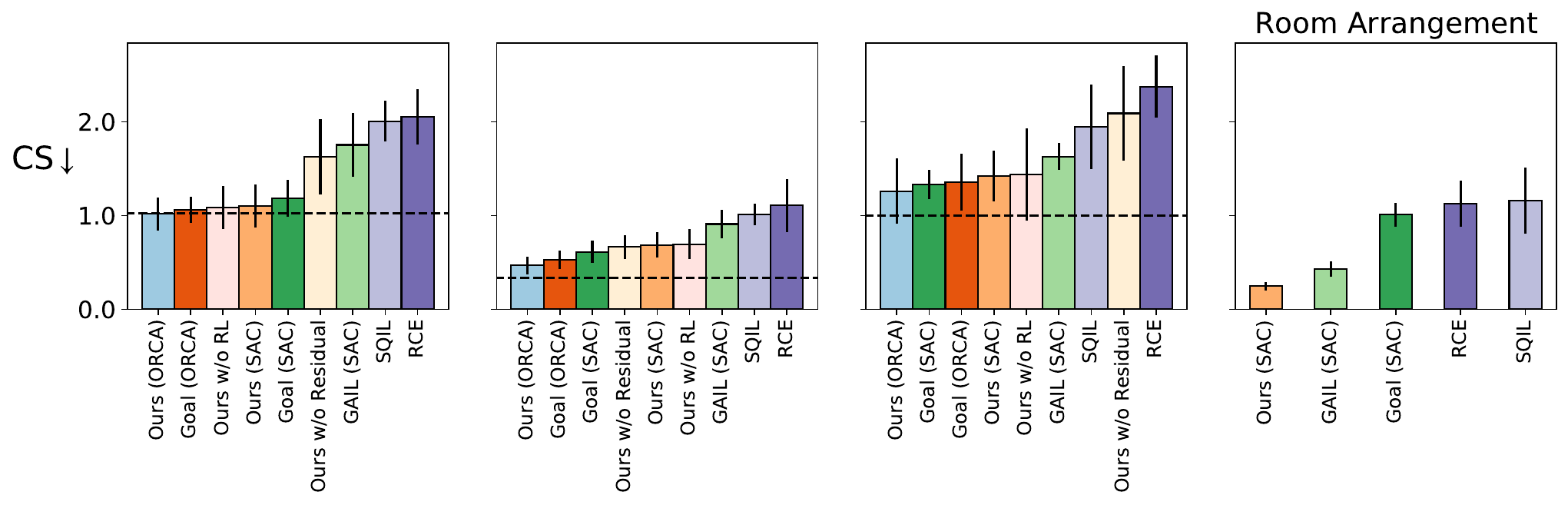}
    \includegraphics[width=\linewidth]{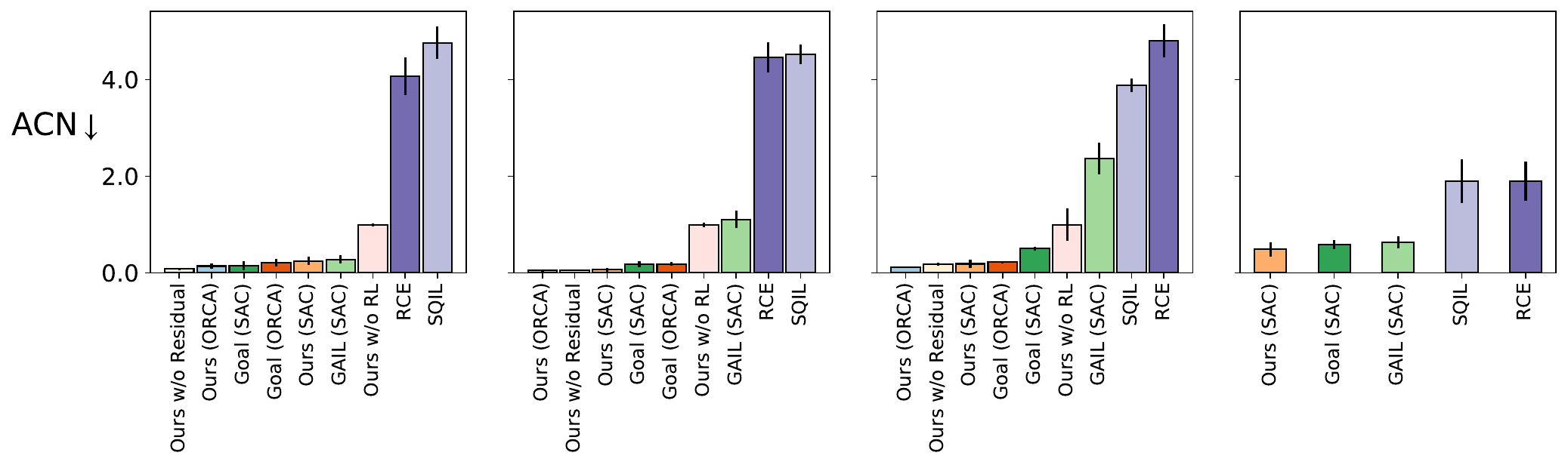}
    \caption{
    Quantitative results on ball rearrangement.
    From top to down, we report pseudo-likelihood (PL), coverage score (CS), and averaged collision number (ACN).
    The PL curves and CS bars of each task are normalised by the oracle's performance of this task.
    For CS and ACN, we report the mean and confidence interval over five random seeds.
    }
    \label{tab:comparative}
    \vspace{-0.3cm}    
\end{figure}
%%%%%%%%%%%%%%%%%%%%%%%%%%%%%%%%%%%%%%%%%%%%%%%%%%%%%%%%%%%%%%%%%%%%%%%%%%%%%%%%%%%%%%%%

\section{Result Analysis}
\subsection{Baseline Comparison}

\textbf{Ball Rearrangement:} 
We present the quantitative results on ball rearrangement tasks in the first three columns of Fig.~\ref{tab:comparative}.
As shown on the top row of Fig.~\ref{tab:comparative}, the PL curves of \textit{Ours~(SAC)} and \textit{Ours~(ORCA)} are significantly higher and steeper than all the baselines in all three tasks, which shows the effectiveness of our method in arranging objects towards the target distribution efficiently.
In the middle row of Fig.~\ref{tab:comparative}, \textit{Ours~(ORCA)} achieves the best performance in CS across all three tasks. \textit{Goal~(SAC)} outperforms \textit{Ours~(SAC)} on \textit{Clustering} and \textit{Circling+Clustering} since the goal is proposed by a generative model that can naturally generate diverse and realistic goals.
From the qualitative results shown in Fig.~\ref{fig:ball qualitative}, our methods, especially \textit{Ours~(ORCA)}, also outperform the baselines in realism. 
Our methods also achieve the lowest ACN in most cases, except that \textit{Goal~(SAC)} is slightly better than \textit{Ours~(SAC)} in \textit{Circling}. 

The \textit{Goal~(SAC)} is the strongest baseline yet loses to our framework in efficiency since the generated goals are far from the initial state.
We validate this problem by comparing the averaged path length of the \textit{Goal~(SAC)} with ours in Appendix.~\ref{sup: ASC}.
Results show that our methods converge in a shorter path than goal-based baselines across all three tasks.
Besides, the \textit{Goal~(SAC)} suffers from invalid goal proposals.
As shown in Fig.~\ref{fig:ball qualitative}, the \textit{Goal~(SAC)} and \textit{Goal~(ORCA)} can rearrange objects in the right trend, but the objects may block each other's way. 
This is because the goals generated by the VAE are not guaranteed to be valid~(\eg the balls may overlap with each other), so the agent cannot reach the invalid goal.
We demonstrate this problem in Appendix.~\ref{sec: vae_results}.

The classifier-based reward learning method, \textit{GAIL}, \textit{RCE}, and \textit{SQIL}, basically fails in most tasks, as shown in Fig.~\ref{fig:ball qualitative} and Fig.~\ref{tab:comparative}.
This is due to the over-exploitation problem that commonly exists in classifier-based reward learning methods. We validate this in Appendix~\ref{sec: reward learning}.

\textbf{Room Rearrangement:}
We only compare \textit{Ours~(SAC)} with learning-based baselines because it is infeasible to employ the ORCA planner in room rearrangement since all the furniture is non-circular.
As shown in Fig.~\ref{fig:room qualitative}, our method can obtain more realistic rearrangement results than baselines.
As shown in the last column of Fig.~\ref{tab:comparative}, 
the coverage score of \textit{Ours~(SAC)} is the lowest and almost half of that of \textit{Goal (SAC)}.
Meanwhile, \textit{Ours~(SAC)} achieves the lowest ACN compared with baselines.
These results prove our method can work on a more complex task and generalise well to unseen conditions.

\subsection{Ablation Studies and Analysis}
We conduct ablation studies on ball rearrangement tasks to investigate: 1) The effectiveness of the \textit{exploration guidance} provided from the target gradient field and 2) The necessity of combining the target gradient field with \textit{control algorithms} to adapt to the environment dynamics.
To this end, we construct an ablated version named \textit{Ours w/o Residual} that drops the residual policy learning and another named \textit{Ours w/o RL} that only takes gradient-based action $\ac^g$ as policy.

\begin{figure}[tb]
    \centering
    % \resizebox{\columnwidth}{}{
    % \scalebox{1}{
        \includegraphics[width=\columnwidth]{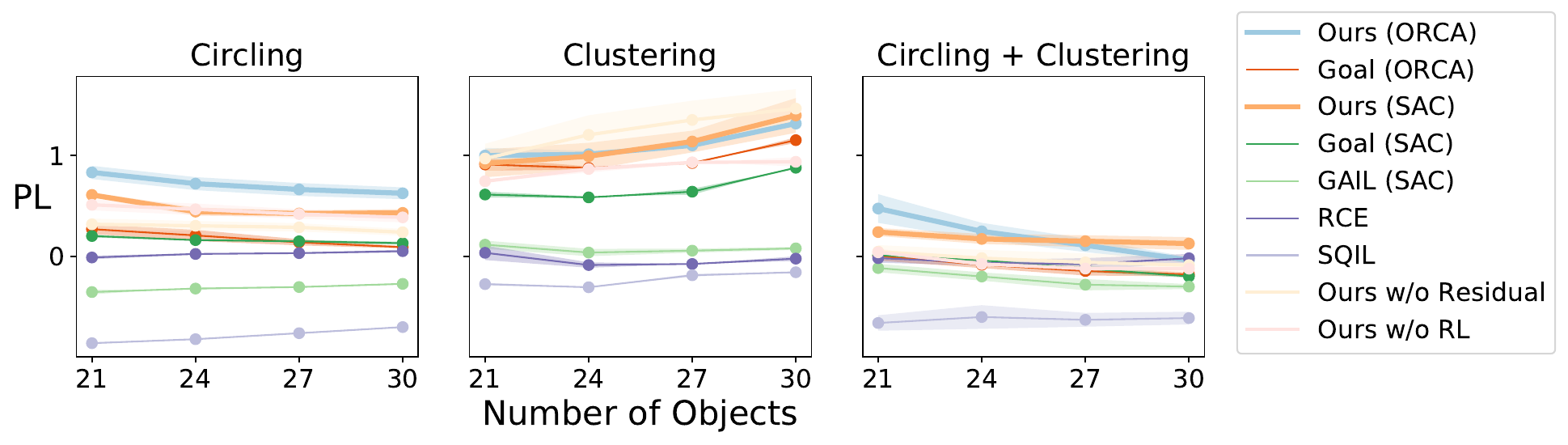}
    % }}
    \vspace{-0.3cm} 
    \label{tab:ball diff num}
    \caption{
    \textbf{Scalability Anaslysis}: Each policy is trained on environments with $3\times 7$ balls and is zero-shot transferred to environments with different ball numbers. 
    }
    \vspace{-0.4cm}
\end{figure}

\textbf{Ours w/o Residual:} The CS bars in Fig.~\ref{tab:comparative} shows that \textit{Ours~(SAC)} significantly outperforms the \textit{Ours w/o Residual} in CS, which means the agent is easy to over-exploit few modes with high density. 
We also show qualitative results in Appendix~\ref{sec: single_mode} that the objects are usually arranged into a single pattern without the residual policy learning.

\textbf{Ours w/o RL:} The PL curves in Fig.~\ref{tab:comparative} show that without the control algorithm serving as the backbone, the rearrangement efficiency of \textit{Ours w/o RL} is significantly lower than \textit{Ours~(SAC)} and \textit{Ours~(ORCA)}. 
Meanwhile, the ACN of \textit{Ours w/o RL} is significantly larger than \textit{Ours~(SAC)} and \textit{Ours~(ORCA)}, which indicates the gradient-based action may cause severe collisions without adapting to the environment dynamics.

\textbf{Scalability:} We further evaluate the scalability of our methods by zero-shot transferring the learned model to rearrange different numbers of objects. 
Specifically, we train all the policies in the ball environment with $ 3 \times 7 = 21$ balls.
During the testing phase, we directly transfer each policy to environments with increased ball numbers, i.e., $3\times 8$, $3\times 9$, and $3\times 10$, without any fine-tuning. 
In Fig. \ref{tab:ball diff num}, we report the averaged PL increment from each method's initial to target states.
Results show that our method still outperforms baselines under different numbers of balls.
% while achieving a slight performance drop.

\section{Conclusion}
\label{sec:conclusion}

In this study, we first analyse the challenges of object rearrangement without explicit goal specification and then incorporate a target gradient field with either model-based planning or model-free learning approaches to achieve this task.
% To tackle this problem, we are the first to introduce the score-based method for object arrangement tasks, where our method can
Experiments demonstrate our effectiveness by comparisons with learning-based and planning-based baselines.

\textbf{Limitation and Future Works.} This work only considers the planar state-based rearrangement with simplified environments where the agent can directly control the velocities of objects.
In future works, it is necessary to extend our framework to more realistic scenarios, \eg, manipulating objects by robots in a realistic 3D environment~\cite{zuo2019craves, qiu2017unrealcv}. We summarise the future directions in the following: 
\begin{itemize}
    \item \textit{Visual Observation: } 
We may explore how to conduct efficient and accurate score-based reward estimation from image-based examples and leverage the recent progress on visual state representation learning~\cite{worldmodel1, worldmodel2, worldmodel3, zhong2019ad} to develop the visual gradient-based action.
    \item \textit{Complex Dynamics: }
When the agent can only move one object at a time, we may explore hierarchical methods where high-level policy chooses which object to move and low-level policy leverages the gradient-based action for motion planning.
When we can only impose forces on objects instead of velocities, we may explore a hierarchical policy where a high-level policy outputs target velocities and a low-level employ a velocity controller, such as a PID controller, to mitigate the velocity errors.
    \item \textit{Multi-Agent Scenarios: } The framework can also be extended to multi-agent scenarios~\cite{zhong2021towards, wang2022tomc, xu2020learning, pan2022mate}, where the decentralised agents need to estimate its gradient field and take actions according to the local observation for multi-agent cooperation or competition.
\end{itemize}

\textbf{Ethics Statement.}\label{ethic} Our method has the potential to build home-assistant robots.
We evaluate our method in simulated environments, which may introduce data bias.
However, similar studies also have such general concerns. We do not see any possible major harm in our study.

\section*{Acknowledgement}
This work was supported by the National Natural Science Foundation of China -Youth Science Fund (No.62006006): Learning Visual Prediction of Interactive Physical Scenes using Unlabelled Videos and China National Post-
doctoral Program for Innovative Talents (Grant No. BX2021008). We also thank Yizhou Wang, Tianhao Wu and Yali Du for their insightful discussions. 

\bibliography{main}
\bibliographystyle{unsrt}

\clearpage

%%%%%%%%%%%%%%%%%%%%%%%%%%%%%%%%%%%%%%%%%%%%%%%%%%%%%%%%%%%%

\section*{Checklist}
\begin{enumerate}

\item For all authors...
\begin{enumerate}
  \item Do the main claims made in the abstract and introduction accurately reflect the paper's contributions and scope?
    \answerYes{} see Sec.~\ref{sec:introduction}
  \item Did you describe the limitations of your work?
    \answerYes{} see Sec.~\ref{sec:conclusion}
  \item Did you discuss any potential negative societal impacts of your work?
    \answerYes{} see supplementary
  \item Have you read the ethics review guidelines and ensured that your paper conforms to them?
    \answerYes{} see Sec.~\ref{sec:conclusion} and supplementary
\end{enumerate}

\item If you are including theoretical results...
\begin{enumerate}
  \item Did you state the full set of assumptions of all theoretical results?
    \answerNA{}
        \item Did you include complete proofs of all theoretical results?
    \answerNA{}
\end{enumerate}

\item If you ran experiments...
\begin{enumerate}
  \item Did you include the code, data, and instructions needed to reproduce the main experimental results (either in the supplemental material or as a URL)?
    \answerYes{} please see the supplemental material.
  \item Did you specify all the training details (e.g., data splits, hyperparameters, how they were chosen)?
    \answerYes{} more details in the supplemental material.
        \item Did you report error bars (e.g., with respect to the random seed after running experiments multiple times)?
    \answerYes{} more details in the supplemental material.
        \item Did you include the total amount of compute and the type of resources used (e.g., type of GPUs, internal cluster, or cloud provider)?
    \answerYes{} more details in the supplemental material.
\end{enumerate}

\item If you are using existing assets (e.g., code, data, models) or curating/releasing new assets...
\begin{enumerate}
  \item If your work uses existing assets, did you cite the creators?
    \answerYes{}
  \item Did you mention the license of the assets?
    \answerNA{}
  \item Did you include any new assets either in the supplemental material or as a URL?
    \answerYes{}
  \item Did you discuss whether and how consent was obtained from people whose data you're using/curating?
    \answerNA{}
  \item Did you discuss whether the data you are using/curating contains personally identifiable information or offensive content?
    \answerYes{}
\end{enumerate}

\item If you used crowdsourcing or conducted research with human subjects...
\begin{enumerate}
  \item Did you include the full text of instructions given to participants and screenshots, if applicable?
    \answerNA{}
  \item Did you describe any potential participant risks, with links to Institutional Review Board (IRB) approvals, if applicable?
    \answerNA{}
  \item Did you include the estimated hourly wage paid to participants and the total amount spent on participant compensation?
    \answerNA{}
\end{enumerate}
\end{enumerate}

%%%%%%%%%%%%%%%%%%%%%%%%%%%%%%%%%%%%%%%%%%%%%%%%%%%%%%%%%%%%

\appendix

\UseRawInputEncoding
\clearpage

\section{Details of Rearrangement Tasks}
\label{sec: rearrange details}
\subsection{Ball Rearrangement}
The three tasks in ball rearrangement only differ in state representation, target distribution and pseudo-likelihood function:

\textbf{Circling:} 
The state is represented as a fully-connected graph where each node contains the two-dimensional position of each ball $\s^i = [x, y]$. 
The target examples are drawn from an explicit process with an intractable density function: 
We first uniformly sample a legal centre $o = [x_o, y_o]$ that allows all balls to place in a circle around this centre without overlap. 
Then we uniformly sample a radius $r_o$ from the legal radius range based on the sampled centre.
After sampling the centre and radius, we uniformly place all balls in a circle centred in $o$ with radius $r_o$.
The pseudo-likelihood function is defined as $\fprox(\s) = \exp^{-(\sigma_{\theta} + \sigma_{r})}$, where $\sigma_{\theta}$ and $\sigma_{r}$ denote the standard deviation of the angle between two adjacent balls and the distances from each ball to the centre of gravity of all balls, respectively. 
Intuitively, if a set of balls are arranged into a circle, then the $\sigma_{r}$ and $\sigma_{\theta}$ should be close to zero, achieving higher pseudo-likelihood.

\textbf{Clustering}:
The state is represented as a full-connected graph where each node contains the two-dimensional position of each ball $\s^i = [x, y]$ and the one-dimensional category feature of each ball $c^i \in \{0, 1, 2\}$.
We generate the target examples in two stages:
First, we sample the positions of each ball from a \textit{Gaussian Mixture Model(GMM)} $p_{GMM}: \R^{K*2} \rightarrow \R^{+}$ with two modes:
\begin{equation}
\begin{aligned}
p_{GMM}(\s) = \frac{1}{2}&\prod_{1\leq i \leq \frac{K}{3}} \N((0.18\sin(\frac{0}{3}\pi), 0.18\cos(\frac{0}{3}\pi)), 0.05I)(\s^i) \\
&\prod_{\frac{K}{3}\leq i \leq \frac{2K}{3}} \N((0.18\sin(\frac{1}{3}\pi), 0.18\cos(\frac{1}{3}\pi)), 0.05I)(\s^i) \\
&\prod_{\frac{2K}{3}\leq i \leq K} \N((0.18\sin(\frac{2}{3}\pi), 0.18\cos(\frac{2}{3}\pi)), 0.05I)(\s^i) \\
+\frac{1}{2}&\prod_{1\leq i \leq \frac{K}{3}} \N((0.18\sin(\frac{0}{3}\pi), 0.18\cos(\frac{0}{3}\pi)), 0.05I)(\s^i) \\
&\prod_{\frac{K}{3}\leq i \leq \frac{2K}{3}} \N((0.18\sin(\frac{2}{3}\pi), 0.18\cos(\frac{2}{3}\pi)), 0.05I)(\s^i) \\
&\prod_{\frac{2K}{3}\leq i \leq K} \N((0.18\sin(\frac{1}{3}\pi), 0.18\cos(\frac{1}{3}\pi)), 0.05I)(\s^i)
\end{aligned}
\end{equation}
where $K=21$ denotes the total number of balls.
In the second stage, we slightly adjust the positions sampled from the GMM to eliminate overlaps between these positions by stepping the physical simulation.
Since the results mainly depend on the GMM, we take the density function of GMM as the pseudo-likelihood function $\fprox(\s) = p_{GMM}(\s)$

\textbf{Circling+Clustering}: 
The state representation is the same as that of \textit{Clustering}. 
To generate the target examples, we first generate a circle using the same process in  \textit{Circling}. 
Then, starting with any ball in the circle,  we colour the ball three colours, in turn, $\frac{K}{3}$ for each.
The order is randomly selected from red-yellow-blue and red-blue-yellow.
This sampling is also explicit yet with an intractable density similar to \textit{Circling}.
The pseudo-likelihood function is defined as $\fprox(\s) = \exp^{-(\sigma_{\theta} + \sigma_{r})} \cdot \exp^{-(\sigma_{R} + \sigma_{G} + \sigma_{B}) - \sigma_{C}}$
where $\sigma_{R}$ denotes the standard deviation of the angle between two adjacent red balls, and $\sigma_{G}$ and $\sigma_{B}$ and $\sigma_{C}$ denotes the standard deviation of the positions of red, green and blue centres.
Intuitively, the first term $\exp^{-(\sigma_{\theta} + \sigma_{r})}$ measures the pseudo-likelihood of balls forming a circle and the next term $\exp^{-(\sigma_{R} + \sigma_{G} + \sigma_{B}) + \sigma_{C}}$ measures the pseudo-likelihood of balls being clustered into three piles.

The common settings shared by each task are summarised as follows:

\textbf{Horizon:} Each training episode contains 100 steps.

\textbf{Initial Distribution:}
We first uniformly sample rough locations for each ball, and then we eliminate overlaps between these positions by stepping physical simulation.

\textbf{Dynamics:}
The floor and wall are all absolutely smooth planes.
All the balls are bounded in an 0.3m x 0.3m area, with a radius of 0.025m.
We set the friction coefficients of all balls to 100.0 since we observe that setting a small(\eg not larger than 1.0) friction coefficient does not significantly affect the dynamics.
Besides, to increase the complexity of the dynamics, we set the masses and restitution coefficients of all green and blue balls to 0.1 and 0.99, respectively.
All red balls' masses and restitution coefficients are set to 10 and 0.1, respectively.
We observe that under these dynamics, the collision may significantly harm the efficiency of the rearrangement process.
Hence, the agent has to adapt to the dynamics for more efficient object rearrangement.

\textbf{Target Examples:} 
We collect 100,000 examples for each task as target examples. 

\subsection{Room Rearrangement}
\textbf{Dataset and Simulator:} 
We clean the 3D-Front dataset~\cite{3dfront} to obtain \textit{bedrooms} of rectangular shape and three to eight objects.
We also drop the rooms with large objects or small free spaces.
Since the number of different types of objects varies greatly, we drop the rooms with objects of rare category~(\eg, top cabinet).
Each room is augmented by flipping two times and rotating four times to get eight variants.
We import these rooms into iGibson~\cite{igibson} to run the physical simulation.
For a more efficient environment reset and physical simulation, we build a `proxy simulator' based on PyBullet~\cite{pybullet} to replace the original iGibson simulator.
We use iGibson to load and save the metadata of each room.
Then we reload these rooms in the proxy simulator, where each object is replaced by a simple box-shaped object with the same geometry.

The 756x8 rooms are used for \textit{target examples}.
The target examples are used for training the target score network, the classifier-based baselines and the VAE in goal-conditioned baselines.
The other 83x8 rooms are used to initialise the room in the test phase:
We first sample a room from the test split and then perform 1000 Brownian steps to obtain the initial state.

\textbf{State and Action Spaces:} The state consists of a aspect ratio $r_a \in \R^+$ and an object state $\s_o \in \R^{K \times 6}$ where $K$ denotes the number of objects. The aspect ratio indicates the shape of the rectangular room $r_a = \text{tanh}(\frac{b_x}{b_y})$ where $b_x$ and $b_y$ denotes the horizontal and vertical wall bounds. The object state is the concatenation of sub-states of all the objects $\s_o = [\s^1, \s^2, ... \s^i, ..., \s^K]$ where the sub-state of the i-th object $\s^i \in \R^6$ consists of 2-D position, 1-D orientation, 2-D bounding box and a 1-D category label.
The action is also a concatenation of sub-actions of all the objects $\ac_o = [\ac^1, \ac^2, ... \ac^i, ..., \ac^K]$. For the i-th object, the action $\ac^i \in \R^3$ consists of a 2-D linear and a 1-D angular velocity.
The whole action space is normalised into a $3\times K$ dimensional unit-box $[-1, 1]^{3\times K}$ by the velocity bounds.

\textbf{Horizon:} Each training episode contains 250 steps.

\textbf{Initial Distribution:}
To guarantee the initial state is accessible to the high-density region of the target distribution, we sample an initial state in two stages:
First, we sample a room from the 83x8 rooms in the test dataset.
Then we perturb this room by 1000 Brownian steps.

\textbf{Dynamics:}
We set the friction coefficient of all the objects in the room to zero, as the room's dynamics are complex enough.

\section{Details of Our Method}
\label{sup: ours detail}
\subsection{Training the Target gradient Field}
\textbf{Complete training objective:} The complete training objective is the SDE-based score-matching objective proposed by~\cite{SDEScoreMatching}:
\begin{equation}
\E_{t \sim \mathcal U(0, 1)}\E_{\s(0) \sim \tar(\s)}\E_{\s(t) \sim p_{0t}(\s(t) \mid \s(0))}[ \tarS(\s(t), t) - \nabla_{\s(t)}\log p_{0t}(\s(t) \mid \s(0))\|_2^2].
\end{equation}
where $p_{0t}(\s(t) \mid \s(0)) = \mathcal{N}(\s(t); \s(0), \frac{1}{2\log \sigma}(\sigma^{2t} - 1) \mathbf{I})$ and $\sigma=25$ is a hyper-parameter.

Using this objective, we can obtain the estimated score \emph{w.r.t.} different levels of the noise-perturbed target distribution $\tar^{t}(\s(t)) = \int p_{0t}(\s(t) \mid \s(0))\tar(\s(0))d\s(0)$ simultaneously.
This way, we can efficiently try different noise levels for efficient hyperparameter tuning.
The choice of $t$ will be described in Sec.~\ref{sec: training detail}.

\textbf{Network Architecture:} 
All the networks~(\eg, score network, actor and critic networks used in learning-based methods) used in our work are designed into three stages, \textit{pre-processing stage}, \textit{message passing stage} and  \textit{output stage}.
The \textit{pre-processing stage} aims at extracting node features for each object via linear layers to construct a fully connected graph for the next stage.
The \textit{message passing stage} takes the initial graph as input and passes them through several graph convolutional layers.
After the above two stages, \textit{output stage} further encodes the feature of each node to obtain the node-wise output~(\eg, score, action). We denote the hidden dimension as $d_h$ and the embedding dimension as $d_e$. 
In room rearrangement, $d_h=128, d_e=64$ and in ball rearrangement $d_h=64, d_e=32$.
We recommend looking up the other trivial details in our open-sourced codes.

We first encode the static feature $f_s^i \in \R^{d_e}$  and state feature $f_a^i \in \R^{d_h}$ for the i-th node  with 2 linear layers.
We further encode the noise feature $f_t \in \R^{d_e}$ by a Gaussian Fourier projection layer~(used in~\cite{SDEScoreMatching}).
For room rearrangement, we additionally encode the wall feature $f_w \in \R^{d_e}$ from the aspect ratio via two linear layers.

Then we construct a fully connected input graph where the content of the i-th node is the concatenation of static, state, noise and wall feature $[f_s^i, f_a^i, f_t, f_a] \in \R^{3*d_e+d_h}$ (for room rearrangement, $[f_s^i, f_a^i, f_t, f_a, f_w] \in \R^{4*d_e+d_h}$).

The input graph is passed through 3~(2 for room rearrangement) Edge Convolutional Layers where the inner network is two layers of MLPs with hidden size $d_h$.
After the message passing, the 2~(3 for room rearrangement) dimensional node features serve as the score components on objects.

Following the parameterisation trick proposed by ~\cite{song2020improved}, we divide the score components by $\frac{1}{2\log \sigma}(\sigma^{2t} - 1)$.

\subsection{Details of ORCA and Planning-based Framework}
\label{sec: orca_detail}
We set $\tau=0.1$ and the simulation duration of each timestep $\Delta t=0.02$.
For each agent(object), ORCA only considers the 2-nearest agents as neighbours since we observe that ORCA often has no solution when the number of neighbours is larger than 2.

In all (ball)~rearrangement tasks, we choose $t=0.1$ as the initial noise scale for the target score network. The noise level linearly decays to 0 within an episode.

\subsection{Complete Derivation of Surrogate Objective}
\begin{equation}
\begin{aligned}
J(\pi)\iff&\E_{ \rho(\s_0), \tau \sim \pi} [ \sum\limits_{\s_t \in \tau} \gamma^t \log \tar(\s_t) ] - \underbrace{\frac{\E_{ \rho(\s_0)} [\log \tar(\s_0)]}{1-\gamma}}_{constant \quad } \\ 
=&\E_{ \rho(\s_0), \tau \sim \pi} [ \sum\limits_{\s_t \in \tau} \gamma^t \log \tar(\s_t) ] - \E_{ \rho(\s_0), \tau \sim \pi} [ \sum\limits_{\s_t \in \tau} \gamma^t \log \tar(\s_0)] \\
=&\E_{ \rho(\s_0), \tau \sim \pi} [\sum\limits_{\s_t \in \tau}\gamma^t[ \log \tar(\s_t) -  \log \tar(\s_0) ]]  \\
=&\E_{ \rho(\s_0), \tau \sim \pi} [\sum\limits_{1 \leq t \leq T}\gamma^t \sum\limits_{1\leq k \leq t}[\log \tar(\s_k) - \log \tar(\s_{k-1}) ]] \overset{def}{=} J^*(\pi) \\
J^*(\pi)\approx& \E_{ \rho(\s_0), \tau \sim \pi} [\sum\limits_{1 \leq t \leq T} \gamma^t \sum\limits_{1\leq k \leq t} \langle \nabla_{\s}\log p(\s_{k-1}), \s_{k}-\s_{k-1} \rangle] \\
\approx&\E_{ \rho(\s_0), \tau \sim \pi}[\sum\limits_{1 \leq t \leq T} \gamma^t \underbrace{\sum\limits_{1\leq k \leq t}\langle \tarS(\s_{k-1}), \s_{k}-\s_{k-1} \rangle}_{r_{t-1}}] \overset{def}{=} \hat{J}(\pi)
\end{aligned}
\end{equation}

\subsection{Details of Learning-based Framework}
\label{sec: training detail}
\textbf{Reward Function:} For all the ball rearrangement tasks, we choose $t=0.1$ for the target score network when outputting the gradient-based action $\ac^g_t = \grad(\tarS(\s_t, 0.1))$ and $t=0.01$ when estimating the immediate reward $r_t = \langle \tarS(\s_{t}, 0.01), \s_{t+1}-\s_{t} \rangle$.
Our experiments found that the choice of $t=0.01$ for the gradient-based action works better. For a fair comparison with Ours(ORCA), we still set $t=0.1$ for the gradient-based action.
At each time step $t$, each object also receives a collision penalty $c^i_t$.
So the total reward for the i-th object at timestep $t$ is $r_t^i = \langle \tarS(\s_{t}, 0.01), \s_{t+1}-\s_{t} \rangle + \lambda * c_t^i$ where $\lambda$ denotes a hyper-parameter to balance the immediate reward and the collision penalty.
We choose $\lambda=5$ for \textit{Clustering} and \textit{Circling+Clustering}, $\lambda=3$ for \textit{Circling} and $\lambda=0.2$ for room rearrangement.
We conduct reward normalisation for both immediate reward and the collision penalty, which maintains a running mean $\mu_t$ and a standard deviation$\sigma_t$ of a given reward sequence and 
returns a z-score $z_t = \frac{r_t - \mu_t}{\sigma_t}$ as the normalised reward. For room rearrangement, we choose $t=0.01$ to output the gradient-based action and estimate the immediate reward.

\textbf{RL Backbone:}
We use Soft-Actor-Critic~(SAC)~\cite{SAC} as our RL backbone and implement SAC based on an open-sourced PyTorch implementation on GitHub~\cite{pytorch_sac} with 300+ stars.
We keep all the hyperparameters the same except that $\gamma = 0.95$ since the reward signal is dense in our case.
We set training iteration to 500,000 for ball rearrangement and 1,000,000 for room rearrangement.

\textbf{Actor Network:}
Similar to the target score network, we first encode the state feature $f_a^i \in \R^{d_h}$ and static feature $f_s^i \in \R^{d_e}$ for the i-th agent.
We also compute the target gradient on the state $\g = \tarS(\s, t)$.
For room rearrangement, we also additionally encode the wall feature $f_w \in \R^{d_e}$ from the aspect ratio via two linear layers.
 
The content of the i-th node of the input graph is the concatenation of the state feature, static feature and gradient component on the i-th object $[f_s^i, f_a^i, \g_i] \in \R^{d_e+d_h+3}$~(For room rearrangement $[f_s^i, f_a^i, f_w, \g_i] \in \R^{d_e*2+d_h+3}$).
After 1~(2 for room rearrangement) layers of message passing via Edge Convolution where the inner network is two layers of MLPs with hidden size $d_h$, the $2\times2$~($3\times2$ for room rearrangement) dimensional node features serve as the mean and variance of the action distribution of the objects.

\textbf{Critic Network:}
The architecture of the critic network is similar to the actor network.
We additionally encode the action feature $f_{ac}^i \in \R^{d_e}$ from the i-th object's action component and then concatenate the action feature with the graph node content. After the message passing, each node contains a one-dimensional output.
We mean pooling the outputs across all nodes to get the output of the critic network.

\section{Details of Baselines}
\label{sup: baselines}
Here we briefly describe the implementation details of baselines.
We recommend directly searching for more details in the supplementary codes.
\subsection{Goal-based Baselines}
These baselines refer to the \textit{Goal-SAC} and \textit{Goal-ORCA} in experiments.

\textbf{Goal Proposal:} This type of baseline first train a VAE on the target examples and then leverages the trained VAE for the goal proposal.
The VAE is implemented as a GNN, and the model capacity is similar to our target score network for a fair comparison.
We choose $\lambda_{kl} = 0.01$ for \textit{Circling} and \textit{Clustering} and $\lambda_{kl} = 0.02$ for \textit{Circling+Clustering}.

\textbf{Execution:} At the beginning of each episode, the agent first proposes a goal for this episode using the VAE and then reaches the goal via a control algorithm.
In \textit{Goal-ORCA}, the agent reaches the goal by a planning-based method:
The agent first assigns velocities for each ball that points to the corresponding goal. 
Then these velocities are updated by the ORCA planner $\planner$ to be collision-free.
In \textit{Goal-SAC}, the agent trains a multi-agent goal-conditioned policy via goal-conditioned RL to reach the goal:
Similar to \textit{Ours-SAC}, the reward of the i-th object at timestep $t$ is $r_t^i = ||\s_t - \s_{goal}||_1 + \lambda * c_t^i$ where $\s_{goal}$ denotes the goal proposal, $\lambda$ is a hyper-parameter and $c_t^i = \sum_{j\neq i} col_{i, j}$ is the total number of collisions between the i-th object and the others.
Here $col_{i, j}=1$ when 
We choose $\lambda=3$ for ball rearrangement and $\lambda=0.2$ for room rearrangement.

\subsection{Classifier-based Baselines}
These baselines refer to the \textit{RCE}, \textit{SQIL} and \textit{GAIL} in experiments.

\textit{RCE} and \textit{SQIL} are implemented based on the codes~\cite{RCE} released by RCE's authors.
We only modify $\gamma = 0.95$, the training steps decrease to 0.5 million for ball rearrangement (\ie, the same number of training steps as other methods) and the model architecture.
The architecture of actor and critic networks is implemented the same as ours(\ie, the same feature extraction layers and Edge Convolutional layers, except for the target gradient feature).

\textit{GAIL} is actually a modification of our learning-based framework:
Keeping the RL agent the same(\ie, multi-agent SAC), \textit{GAIL}'s reward is given by a discriminator.
The architecture of the discriminator is the same as our critic network, except that the input graph does not contain the action feature.

At each training step, we update the discriminator by distinguishing between the agents' and the expert's states(for one step) and then update the RL policy under the reward given by the discriminator(for one step).
The agent also receives a collision reward during training similar to \textit{Ours-SAC} and \textit{Goal-SAC}: $r_t^i = D(\s_{t+1}) + \lambda * c_t^i$ where $D$ denotes the classifier trained by \textit{GAIL}.
We do not conduct reward normalisation for \textit{GAIL} as the learned reward is unstable.
We choose $\lambda=10$ for room rearrangement, \textit{Circling} and \textit{Circling+Clustering}, $\lambda=100$ for \textit{Clustering}.

\section{Details of Evaluations}
\label{sec: eval_details}
\subsection{Ball rearrangement}
We collect 100 trajectories for each task starting from the same set of initial states.
To calculate the coverage score, we sample fixed sets examples from the target distribution serving as $S_{gt}$ for \textit{Circling}, \textit{Clustering}, and \textit{Circling+Clustering}, respectively.
We sample 20 examples for \textit{Circling} and \textit{Circling+Clustering} and 50 examples for \textit{Clustering}.
Since the balls in the same category can actually be viewed as a two-dimensional point cloud, we measure the distance between two states by summing the CDs between each pile of balls by category.

\subsection{Room rearrangement}
For each room in 83 test rooms, we collect eight initial states.
Then we collect 83x8 trajectories starting from these initial states for each method.
The coverage score is calculated by averaging the coverage score in each room condition since the state dimension differs in different rooms.
For each room in 83 test rooms, we calculate the coverage score between the eight ground truth states and eight rearrangement results and then the averaged coverage score over the 83 rooms is taken as the final coverage score for a method. We measure the distance between two states by calculating the average L2 distance between the positions~(\ie, we ignore the orientations) of the corresponding objects.

% \section{Additional Ablation Studies}
% \subsection{Full Term vs Last Term}

\section{Additional Results}
\subsection{Single-mode Problem of Ours w/o Residual}
\label{sec: single_mode}

%%%%%%%%%%%%%%%%%%%%%%%%%%% Single Pattern of non-residual %%%%%%%%%%%%%%%%%%%%%%%%%%%%%%%%%
\begin{figure}[b]   
    \centering
    % \resizebox{\columnwidth}{!}{
    % \scalebox{1}{
        \includegraphics[width=1\columnwidth]{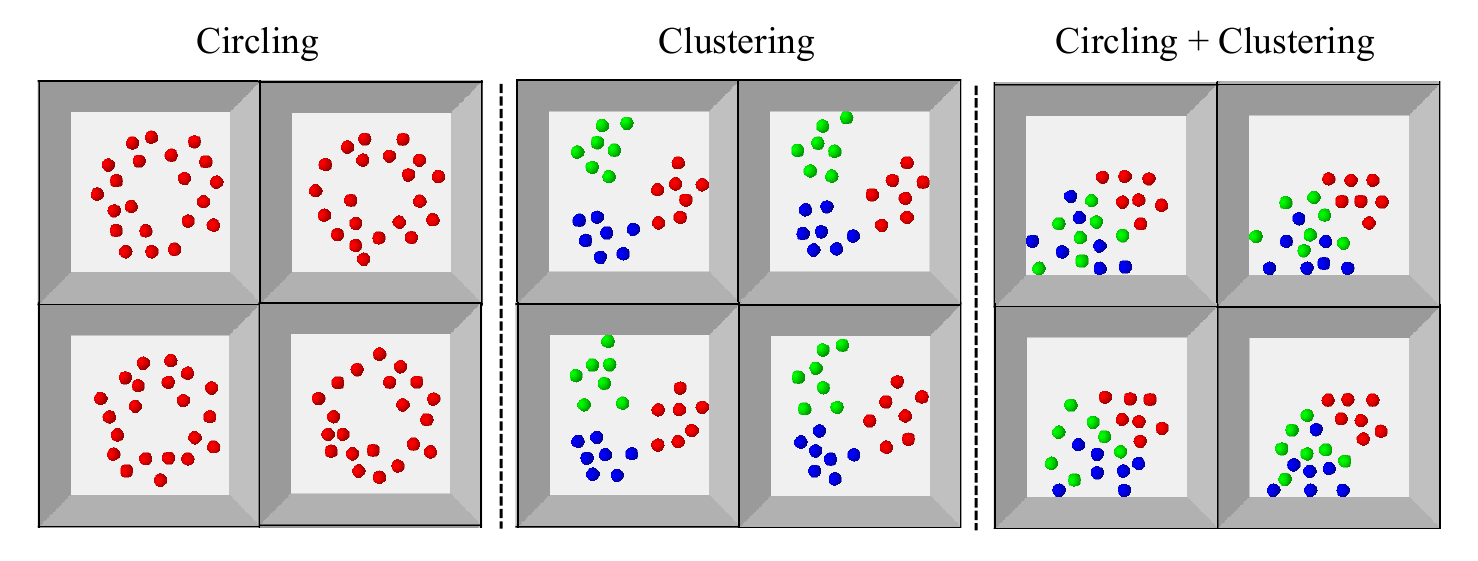}
    % }}
    \caption{
    We visualise rearrangement results of \textit{Ours w/o Residual} to demonstrate the `single pattern' phenomenon.
    }
    \label{fig:visualisation non-resi}
    % \vspace{-4pt}
\end{figure}
%%%%%%%%%%%%%%%%%%%%%%%%%%%%%%%%%%%%%%%%%%%%%%%%%%%%%%%%%%%%%%%%%%%%%%%%%%%%%%%%%%%%%%%%

In Fig.~\ref{fig:visualisation non-resi} we show qualitative results of \textit{Ours w/o Residual}.
Apparently, the balls are arranged into a single pattern in all three ball rearrangement tasks, while the examples from the target distribution are diverse.
In the most difficult task \textit{Circling + Clustering}, the agent cannot even reach a terminal state with a high likelihood.
This result indicates that \textit{Ours w/o Residual} failed to explore the high-density region of target distribution without residual learning.

% \subsection{Over Exploitation of Classifier-based Methods}
\subsection{Effectiveness of Reward Learning}\label{sec: reward learning}
In each ball rearrangement task, we collect 100 trajectories, each of which is run by a hybrid policy.
The hybrid policy takes the first 50 steps using the \textit{Ours~(ORCA)} and the next 50 steps using random actions.
To evaluate the effectiveness of our method on reward learning, we compare the estimated reward curve of \textit{Ours~(SAC)} and \textit{GAIL} with the \textit{pseudo likelihood(PL)} curve of the trajectories.

\begin{figure}[t] 
    \centering
    \begin{minipage}[t]{0.30\columnwidth}
    \includegraphics[width=\columnwidth]{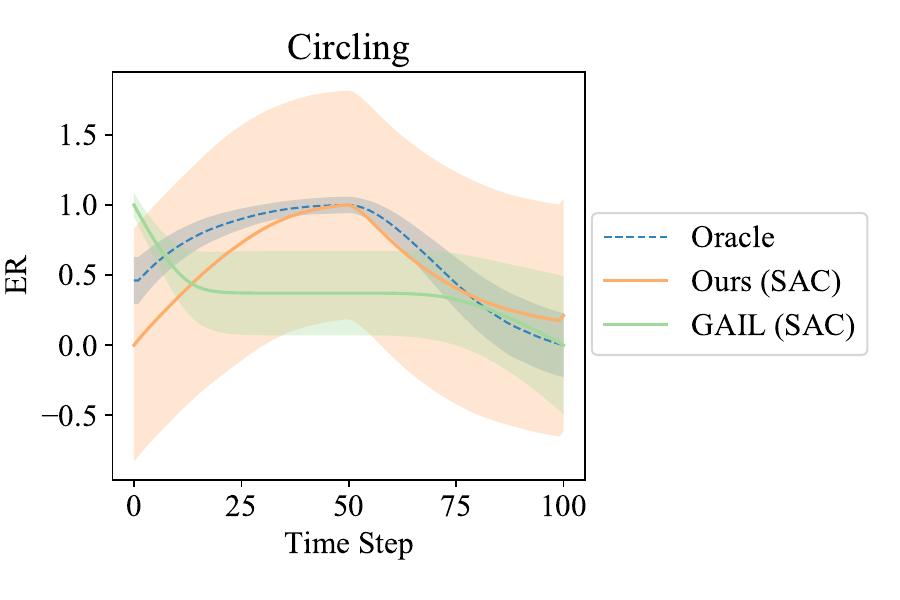}
    \end{minipage}
    \begin{minipage}[t]{0.30\columnwidth}
    \includegraphics[width=\columnwidth]{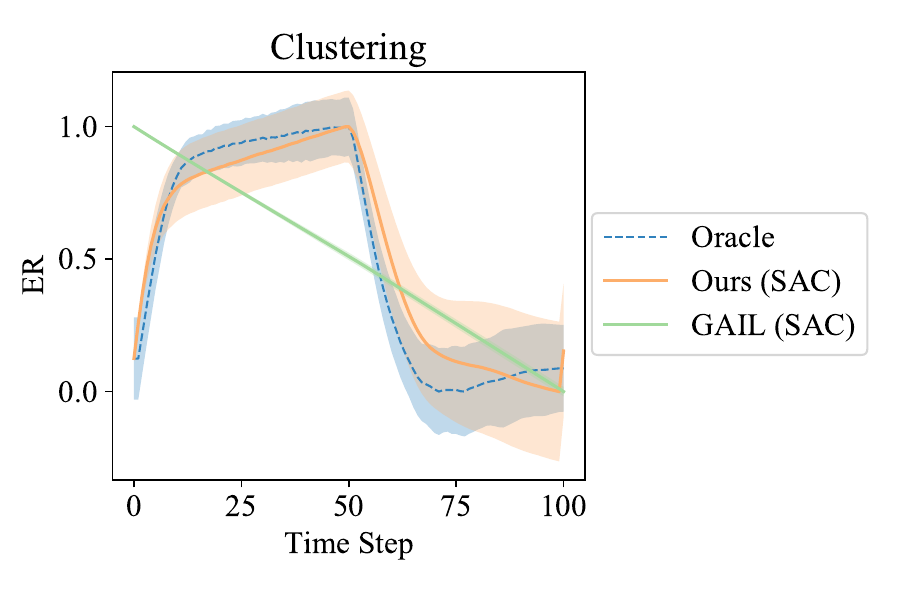}
    \end{minipage}
    \begin{minipage}[t]{0.30\columnwidth}
    \includegraphics[width=\columnwidth]{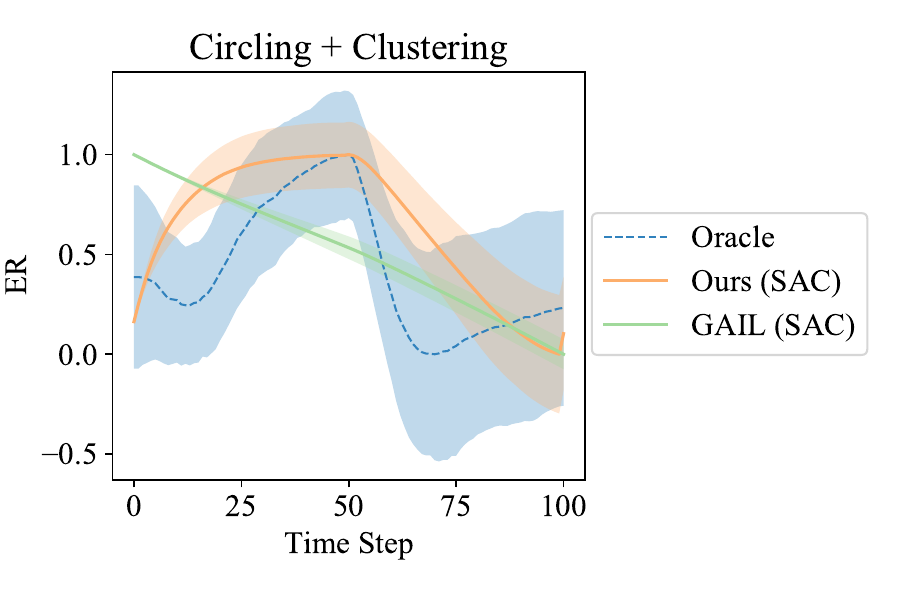}
    \end{minipage}
    
    \caption{
       Given a set of identical trajectories, we compare the \textit{estimated reward(ER)} of different methods and \textit{pseudo likelihood(PL)}.
       All curves are normalised to the range of [- 1,1].
    }
    \label{fig:reward compare}
    % \vspace{-4pt}
\end{figure}

As shown in~\ref{fig:reward compare}, the reward curve of \textit{Ours~(SAC)} best fits the trend of the pseudo-likelihood curve, which shows the effectiveness of our reward estimation method.

%%%%%%%%%%%%%%%%%%%%%%%%%%% Reward Learning Comparison %%%%%%%%%%%%%%%%%%%%%%%%%%%%%%%%%
%%%%%%%%%%%%%%%%%%%%%%%%%%%%%%%%%%%%%%%%%%%%%%%%%%%%%%%%%%%%%%%%%%%%%%%%%%%%%%%%%%%%%%%%

%%%%%%%%%%%%%%%%%%%%%%%%%%% VAE Goals %%%%%%%%%%%%%%%%%%%%%%%%%%%%%%%%%
\begin{figure}[t] 
    \centering
    % \resizebox{\columnwidth}{!}{
    % \scalebox{1}{
        \includegraphics[width=1\columnwidth]{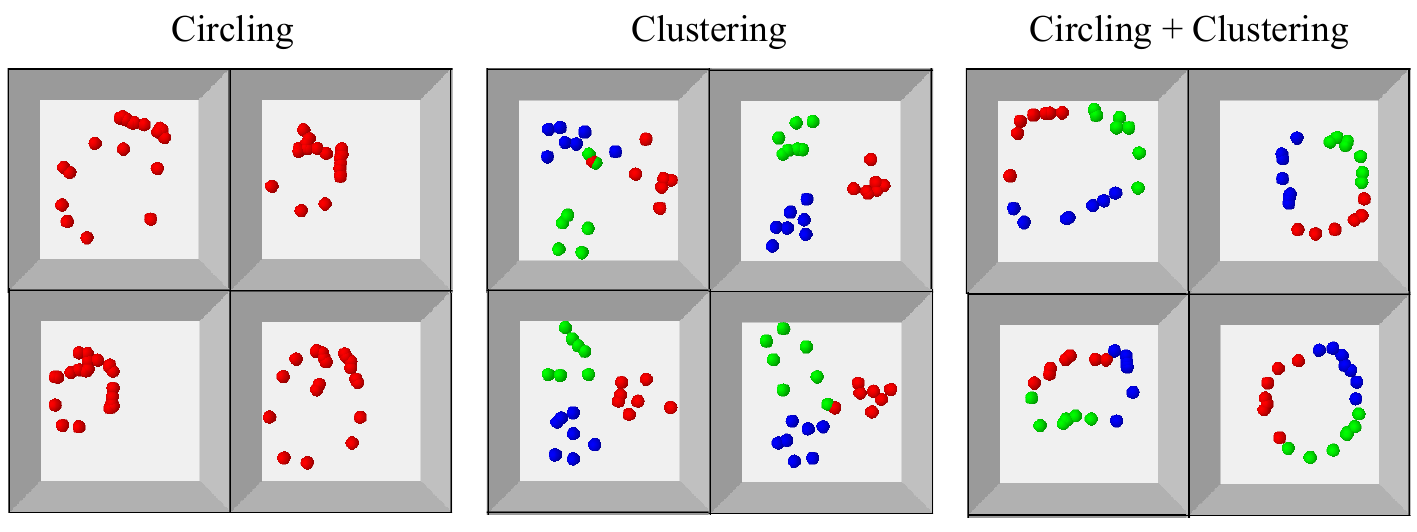}
    % }}
    \caption{
    We visualise the goals proposed by the VAE used in \textit{Goal~(ORCA)} and \textit{Goal~(SAC)}.
    }
    \label{fig:vae goal}
    % \vspace{-4pt}
\end{figure}
%%%%%%%%%%%%%%%%%%%%%%%%%%%%%%%%%%%%%%%%%%%%%%%%%%%%%%%%%%%%%%%%%%%%%%%%%%%%%%%%%%%%%%%%

\subsection{Efficiency Problem of Goal-conditioned Baselines}
\label{sup: ASC}
In Fig.~\ref{tab:ASC}, we report the comparative results of our framework and goal-based baselines(\eg, \textit{Goal~(SAC)}, \textit{Goal~(ORCA)}) on a new metric named \textit{absolute state change(ASC)}.
The ASC measures the sum of the absolute paths of all small balls in the rearrangement process.
\begin{equation}
    \text{ASC} = \sum_{1\leq t \leq T}\sum_{1\leq k \leq K} ||\s^k_t - \s^k_{t-1}||_1
\end{equation}

As shown in Fig.~\ref{tab:ASC}, \textit{Ours~(SAC)} and \textit{Ours~(ORCA)} are significantly better than \textit{Goal~(SAC)} and \textit{Goal~(ORCA)}, respectively.
This result explains why our method's likelihood curves are better than the goal-based baselines': 
The proposed goal is far away from the initial state, which harms the efficiency of goal-based approaches.

\begin{table}[htb]
    \centering
    \vspace{-1mm}    

    \caption{Quantitative comparison results of our framework(\textit{Ours~(ORCA)}, \textit{Ours~(SAC)}) and goal-based baselines(\textit{Goal~(ORCA)}, \textit{Goal~(SAC)}) in rearrangement efficiency. For each method, we report the mean and standard deviation of \textit{absolute state change} over 100 episodes on each ball rearrangement task.}
    \label{tab:ASC}
    \resizebox{\columnwidth}{!}{
    \scalebox{1}{
        \UseRawInputEncoding
\begin{tabular}{cccccccc}
\toprule
\begin{tabular}{ccccccc}
\multirow{2}{*}{Method} & \multicolumn{2}{c}{Circling}                    & \multicolumn{2}{c}{Clustering}                   & \multicolumn{2}{c}{Circling+Clustering}         \\
                        & 21 balls               & 30 balls               & 21 balls               & 30 balls                & 21 balls               & 30 balls               \\
\midrule
Ours-ORCA               & \textbf{23.16 +- 2.42} & \textbf{29.08 +- 2.43} & \textbf{13.72 +- 1.40} & \textbf{16.60 +- 1.68}  & \textbf{19.54 +- 2.19} & \textbf{23.29 +- 2.18} \\
Goal-ORCA               & 28.05 +- 2.82          & 31.98 +- 2.90          & 27.57 +- 2.90          & 33.37 +- 4.06           & 27.77 +- 2.86          & 32.47 +- 2.99          \\
\midrule
Ours-SAC                & \textbf{56.76 +- 4.83} & \textbf{78.18 +- 6.75} & \textbf{65.35 +- 4.64} & \textbf{110.06 +- 4.25} & \textbf{48.93 +- 4.68} & \textbf{80.01 +- 6.19} \\
Goal-SAC                & 108.25 +- 7.53         & 139.26 +- 8.24         & 118.15 +- 5.84         & 142.05 +- 7.56          & 122.72 +- 5.93         & 161.01 +- 6.84         \\
\bottomrule
\end{tabular}
\end{tabular}
    }}
    \vspace{-1mm}    
\end{table}

\subsection{Visualisations of Goals Proposed by the VAE}
\label{sec: vae_results}

We demonstrate the visualisations of goal proposals by the VAE used in goal-based baselines.
Typically, the proposed goals indeed form a reasonable shape that is similar to the target examples.
However, it is hard to generate a fully legal goal since the VAE is not accessible to the dynamics of the environment or has enough data to infer the physical constraints of the environment.
As shown in Fig.~\ref{fig:vae goal}, there exist many overlaps between balls in generated goals, which causes the balls to have conflicting goals and thus harms the efficiency of goal-based baselines.

\subsection{Six-modes Clustering}
\textbf{Task Settings:} 
This task is a six-modes extension of \textit{Clustering}, where the centres of clusters are located in the following six patterns.

\begin{table}[h]
    \centering
    \vspace{-1mm}    
    
    \label{tab:SixModes Setting}
    \resizebox{0.8\columnwidth}{!}{
    \UseRawInputEncoding
\begin{tabular}{c|cccccc}
\toprule
Locations & Mode1 & Mode2 & Mode3 & Mode4 & Mode5 & Mode6 \\
\midrule
$(0.18\cos(\frac{2\pi}{3}), 0.18\sin(\frac{2\pi}{3}))$    & R     & R     & B     & B     & G     & G     \\
$(0.18\cos(\frac{4\pi}{3}), 0.18\sin(\frac{4\pi}{3}))$    & G     & B     & G     & R     & B     & R     \\
$(0.18\cos(2\pi), 0.18\sin(2\pi))$    & B     & G     & R     & G     & R     & B     \\
\bottomrule
\end{tabular}

    }
    % \caption{Bayesian Inference of Latent Distribution}
    \vspace{-1mm}    
\end{table}

Defining the joint centres' positions as a latent variable $C = (C_r, C_g, C_b)$ where $C_r$, $C_g$ and $C_b$ denote centres of red, green and blue balls, respectively and the above six modes as $\{c_i\}_{1 \leq i \leq 6}$, the $C$ obeys a categorical distribution $ p(C=c_i) = \frac{1}{6}$.

The target distribution of six-modes clustering is a Gaussian Mixture Model:

\begin{equation}
\begin{aligned}
p_{GMM}(\s) &= \sum_{1 \leq k \leq 6} p(C=c_k) p(\s|C=c_k ) \\
p(\s|C=c_k ) &=  \prod_{1\leq i \leq \frac{K}{3}} \N(C_r^k, 0.05I)(\s^i) \prod_{\frac{K}{3}\leq i \leq \frac{2K}{3}} \N(C_g^k, 0.05I)(\s^i) \prod_{\frac{2K}{3}\leq i \leq K} \N(C_b^k, 0.05I)(\s^i)
\end{aligned}
\end{equation}

\begin{figure}[htb]
    \centering
    % \resizebox{\columnwidth}{!}{
    % \scalebox{1}{
        \includegraphics[width=1\columnwidth]{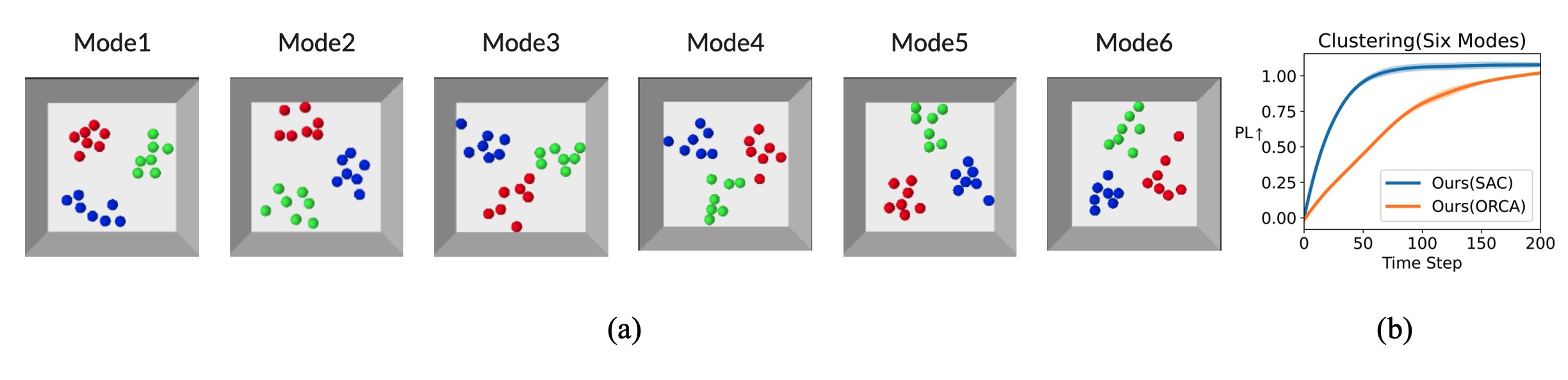}
    % }}
    \caption{
    Pseudo likelihood curves and qualitative results of six-modes clustering.
    }
    \label{fig:sixmodes}
    % \vspace{-4pt}
\end{figure}

Notably, the `mean mode' of the above six modes is the origin, \ie, $\frac{1}{6}\sum_{i=1}^6 c_i = ((0, 0), (0, 0), (0, 0))$.
If the policy arranges the balls into this mean pattern, then the balls should be centred around $(0, 0)$ (\ie, all positions of balls obey $\N(0, 0.05I)$).
As shown in Fig.~\ref{fig:sixmodes} (a), we illustrate an example for each mode.

\textbf{Implementation Details:}
We evaluate Ours~(SAC) and Ours~(ORCA) on this task. The performances are normalised by the Oracle.

\textbf{Results:} 
We report the pseudo-likelihood curves of our methods.
As shown in Fig.~\ref{fig:sixmodes} (b), the rearrangement results are close to ground truth examples from the target distribution.

%%%%%%%%%%%%%%%%%%%%%%%%%%% Six Modes Demo + Curve %%%%%%%%%%%%%%%%%%%%%%%%%%%%%%%%%
%%%%%%%%%%%%%%%%%%%%%%%%%%%%%%%%%%%%%%%%%%%%%%%%%%%%%%%%%%%%%%%%%%%%%%%%%%%%

We further compute the latent distributions for rearrangement results of Ours~(SAC) using the Bayesian Theorem:
\begin{equation}
    p(C = c_i|\s) = \frac{p(\s|C=c_i)p(C=c_i)}{\sum_{j} p(\s|C=c_j)p(C=c_j)} 
\end{equation}
$p(C = c_i|\s)$ indicates which mode a state belongs to. 
To demonstrate our rearrangement results are close to one of the six modes, instead of concentrating on a `mean' pattern, we evaluate the average entropy of the latent distributions $\E[H(p(C|\s))]$. 
As shown in Fig.~\ref{tab:Latent}, the average entropy of our methods is even lower than ground truths'.
This indicates each state in our rearrangement results distinctly belonging to one of the categories.

We also report the average latent distribution $\E[p(C|\s)]$. As shown in Fig.~\ref{tab:Latent}, our methods' averaged latent distribution (overall rearrangement results) achieve comparable orders of magnitude in different modes. This shows the rearrangement of our methods can cover all the mode centres.

\begin{table}[]
    \centering
    \vspace{-1mm}    
    
    \label{tab:Latent}
    \resizebox{\columnwidth}{!}{
    \UseRawInputEncoding
\begin{tabular}{c|c|cccccc}
\toprule
Methods    & \multicolumn{1}{l}{Average Entropy}$\downarrow$ & Mode1 & Mode2 & Mode3 & Mode4 & Mode5 & Mode6 \\
\midrule
GT         & 0.0066                              & 0.17  & 0.17  & 0.17  & 0.17  & 0.17  & 0.17  \\
Ours(SAC)  & 0.0037                              & 0.15  & 0.18  & 0.31  & 0.14  & 0.10  & 0.13  \\
Ours(ORCA) & 0.0056                              & 0.17  & 0.17  & 0.16  & 0.11  & 0.19  & 0.19  \\
\bottomrule
\end{tabular}
    }
    \vspace{-1mm}    
\end{table}

\subsection{Move One-ball at a Time}
\textbf{Task Settings:}
This task is an extension of  \textit{Clustering} where the policy can only move one ball at a time.
We increase the horizon of each episode from 100 to 300.
At each time step, the agent can choose one ball to take a velocity-based action for 0.1 seconds.

\begin{figure}[htb]   
    \centering
    % \resizebox{\columnwidth}{!}{
    % \scalebox{1}{
        \includegraphics[width=1\columnwidth]{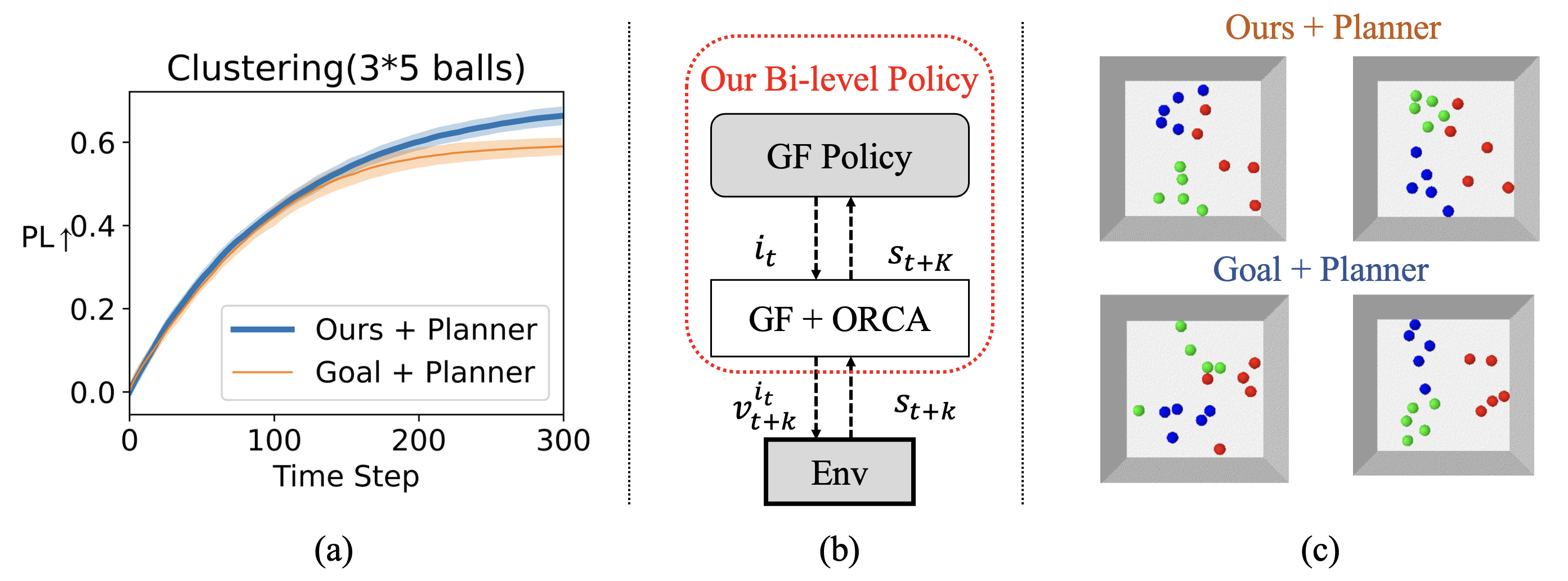}
    % }}
    \caption{
    (a): Quantitative results. (b): Illustration of our bi-level policy. (c): Qualitative results.
    }
    \label{fig:onebyone}
    % \vspace{-4pt}
\end{figure}

\textbf{Implementation Details: }
We design a bi-level approach based on our method as shown in Fig.~\ref{fig:onebyone} (b):
Every 20-time steps, the high-level planner outputs an object index $i_t$ with the largest target gradient's component:
\begin{equation}
    i_t = \underset{i}{\operatorname{argmax}} ||\g_t^i||_2
\end{equation}
where $\g_t^i \in \R^{2}$ denotes the component of the target gradient on the i-th object. 
In the following 20 steps, the ORCA planner computes the target velocity according to $\g_t^i$ and masks all other objects' velocities to zero.

We compare our method with a goal-based baseline where the agent generates goals for each object via the VAE used in Goal~(ORCA). 
Then the high-level planner chooses the object with the farthest distance to the goal, denoted as $i_t$.
The low-level planner of this baseline is the same as ours.

\textbf{Results:}
As shown in Fig.~\ref{fig:onebyone} (a), (c), our method achieves more appealing results, better efficiency and performance compared with the goal-based baseline.

%%%%%%%%%%%%%%%%%%%%%%%%%%% One by One %%%%%%%%%%%%%%%%%%%%%%%%%%%%%%%%%
%%%%%%%%%%%%%%%%%%%%%%%%%%%%%%%%%%%%%%%%%%%%%%%%%%%%%%%%%%%%%%%%%%%%%%%%%%%%%%%%%%%%%%%%

\subsection{Force-based Dynamics}
\textbf{Task Settings:}
This task is an extension of \textit{Circling + Clustering} where the agent can only impose forces on the objects instead of velocities.
At each time step, the agent can assign a two-dimensional force $f_i \in \R^2$ on each object.

\begin{figure}[htb] 
    \centering
    \includegraphics[width=1\columnwidth]{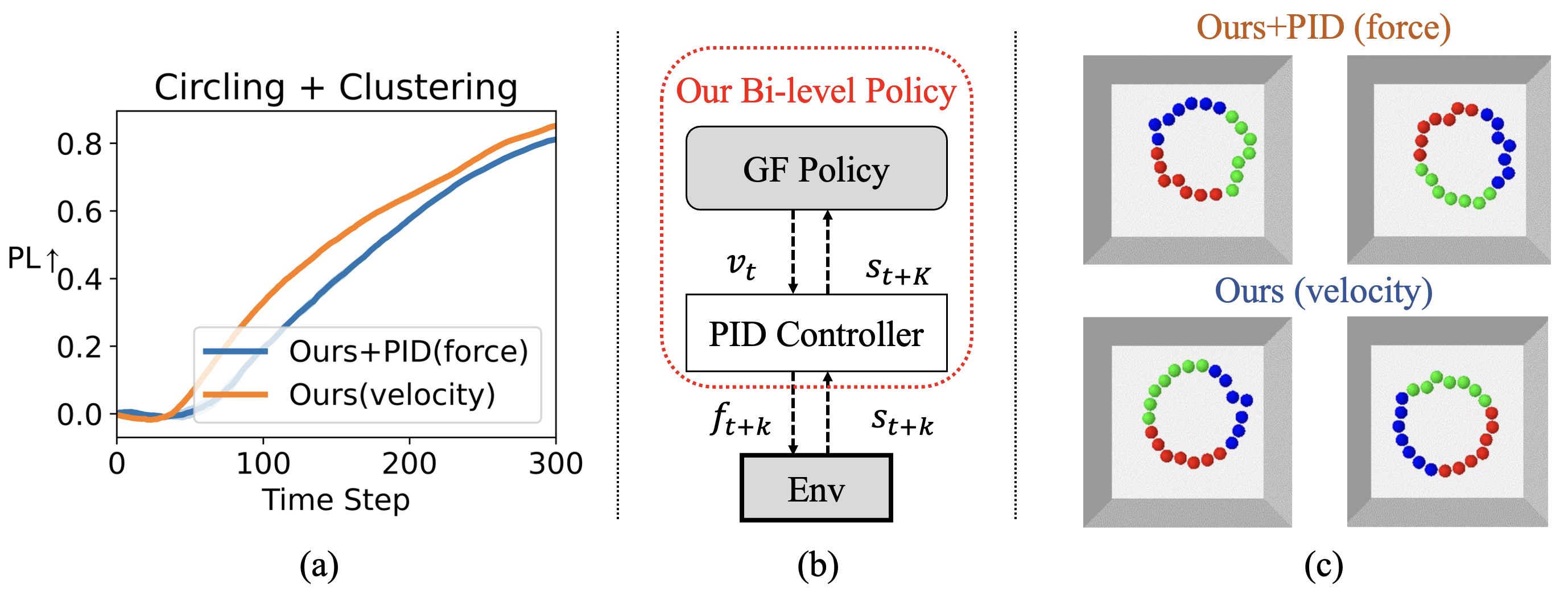}
    \caption{
    (a): Quantitative results. (b): Illustration of our bi-level policy. (c): Qualitative results.
    }
    \label{fig:force}
\end{figure}

\textbf{Implementation Details: }
Similar to the `one at a time' experiment, we design a bi-level approach to tackle this task, as Fig.~\ref{fig:force} (b) illustrates.
The high-level policy outputs a target velocity every eight steps.
In the following eight steps, after the target velocities are outputted, the low-level PID controller receives the target velocity and outputs the force-based action to minimise the velocity error. We set $K_P = 10.0, K_I = 0.0, K_D = 0.0$ for PID controller.

This policy is compared with Ours~(SAC) in the main paper.

\textbf{Results:}
As shown in Fig.~\ref{fig:force}. (a) and (c), this force-based policy achieves comparable performance with Ours~(SAC) yet suffers from a slight efficiency drop due to the control error of PID.

%%%%%%%%%%%%%%%%%%%%%%%%%%% Force Dynamics %%%%%%%%%%%%%%%%%%%%%%%%%%%%%%%%%
%%%%%%%%%%%%%%%%%%%%%%%%%%%%%%%%%%%%%%%%%%%%%%%%%%%%%%%%%%%%%%%%%%%%%%%%%%%%%%%%%%%%%%%%

\subsection{Image-based Reward Learning}
%%%%%%%%%%%%%%%%%%%%%%%%%%% Image Reward %%%%%%%%%%%%%%%%%%%%%%%%%%%%%%%%%

\begin{wrapfigure}{htb}{0.4\textwidth}
\centering
\vspace{-20pt}
\includegraphics[width=0.4\textwidth]{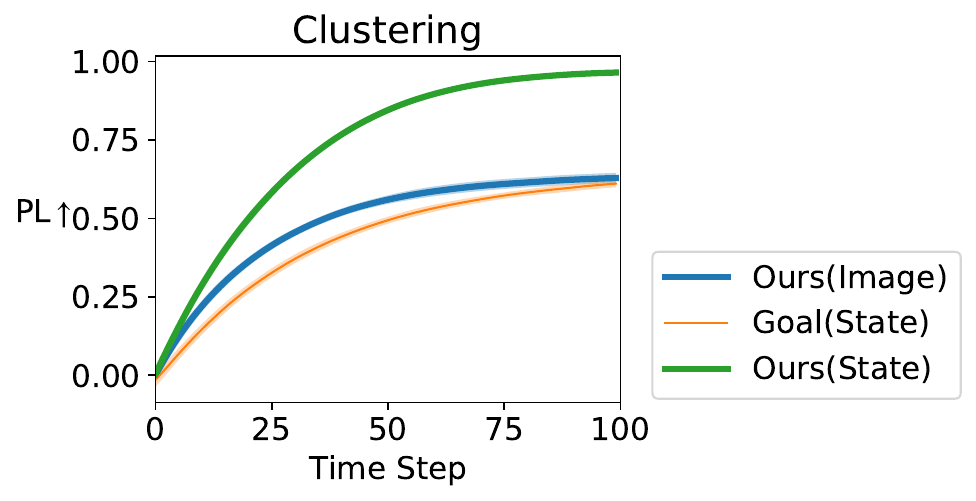}
\caption{Quantitative results of image-based reward learning experiment.}
\label{sec: image-based}
\vspace{-40pt}
\end{wrapfigure}
%%%%%%%%%%%%%%%%%%%%%%%%%%%%%%%%%%%%%%%%%%%%%%%%%%%%%%%%%%%%%%%%%%%%%%%%%%%%%%%%%%%%%%%%
\textbf{Task Settings:}
The target score network is trained on image-based target examples. 
We simply render the state-based target example set used in Ours~(SAC) to 64x64x3 images and 

\textbf{Implementation Details: }
Our target score network of Ours(Image) is trained on the image-based examples set.
The policy network and gradient-based action are state-based since we focus on image-based reward learning instead of visual-based policy learning.

This approach is compared with Ous(SAC) and Goal~(SAC) in the main paper.

\textbf{Results:}
Results in Fig.~\ref{sec: image-based} show that Ours(Image) achieves slightly better performance than Goal~(SAC) yet is lower than Ours~(SAC) due to the increment of the dimension.
This indicates that our reward learning method is still effective in image-based settings.

\end{document}